\documentclass{article}

\usepackage[preprint]{neurips_2025}

\makeatletter
\renewcommand{\@noticestring}{\textit{Proceedings of the
$\mathit{43}^{rd}$ International Conference on Machine Learning},
Seoul, South Korea. PMLR 306, 2026.
Copyright 2026 by the author(s).}
\makeatother

\usepackage[utf8]{inputenc}
\usepackage[T1]{fontenc}
\usepackage{microtype}
\usepackage{graphicx}
\usepackage{subcaption}
\usepackage{booktabs}
\usepackage{hyperref}
\usepackage{url}
\usepackage{amsmath}
\usepackage{amssymb}
\usepackage{amsfonts}
\usepackage{mathtools}
\usepackage{amsthm}
\usepackage[capitalize,noabbrev]{cleveref}
\usepackage{multirow}
\usepackage{makecell}
\usepackage{tabularx}
\usepackage{array}
\usepackage{enumitem}
\usepackage{placeins}
\usepackage{siunitx}
\usepackage{algorithm}
\usepackage{algorithmicx}
\usepackage{algpseudocode}
\usepackage{wrapfig}
\usepackage[most]{tcolorbox}
\usepackage[normalem]{ulem}
\usepackage{xcolor}

\theoremstyle{plain}
\newtheorem{theorem}{Theorem}[section]

\theoremstyle{definition}

\theoremstyle{remark}

\usepackage{amsmath,amsfonts,bm}

\def\1{\bm{1}}

\def\cN{{\mathcal{N}}}

\def\vmu{{\bm{\mu}}}

\def\va{{\bm{a}}}

\def\vu{{\bm{u}}}
\def\vv{{\bm{v}}}

\def\vx{{\bm{x}}}
\def\vy{{\bm{y}}}

\def\vmu{{\bm{\mu}}}

\def\mI{{\bm{I}}}

\DeclareMathAlphabet{\mathsfit}{\encodingdefault}{\sfdefault}{m}{sl}
\SetMathAlphabet{\mathsfit}{bold}{\encodingdefault}{\sfdefault}{bx}{n}

\newcommand{\E}{\mathbb{E}}

\makeatletter
\def\thickhline{%
  \noalign{\ifnum0=`}\fi\hrule \@height \thickarrayrulewidth \futurelet
   \reserved@a\@xthickhline}
\def\@xthickhline{\ifx\reserved@a\thickhline
               \vskip\doublerulesep
               \vskip-\thickarrayrulewidth
             \fi
      \ifnum0=`{\fi}}
\makeatother

\newlength{\thickarrayrulewidth}
\title{Physics-Informed Distillation of Diffusion Models for PDE-Constrained Generation}

\author{%
  Yi Zhang\\
  Institute of Data Science\\
  The University of Hong Kong\\
  \And
  Peng Wang\\
  Department of Computer and Information Science\\
  University of Macau\\
  \And
  Difan Zou\\
  Institute of Data Science \& School of Computing and Data Science\\
  The University of Hong Kong\\
  \texttt{dzou@cs.hku.hk}
}

\begin{document}

\maketitle

\begin{abstract}
Diffusion models show growing promise for generative modeling of physical systems, but enforcing partial differential equation (PDE) constraints directly is infeasible during the stochastic denoising process. Current methods apply constraints to the expected clean sample, incurring a \textit{Jensen's Gap} that forces a trade-off between PDE satisfaction and generative accuracy. To bridge this gap, we propose \textbf{P}hysics-\textbf{I}nformed \textbf{D}istillation of \textbf{D}iffusion \textbf{M}odels (PIDDM), a simple yet effective post-hoc distillation strategy that enforces PDE constraints after training. PIDDM enables fast single-step generation while improving both physical consistency and sample quality, supporting forward/inverse problems and reconstruction from partial observations. Extensive experiments across PDE benchmarks show PIDDM outperforms recent baselines, such as PIDM~\cite{bastek2025pdim}, DiffusionPDE~\cite{huang2024diffusionpde}, and ECI-sampling~\cite{cheng2025eci}, in both accuracy and constraint satisfaction, with lower computation and minimal hyperparameter tuning, offering a more efficient pathway to physics-informed diffusion models.
\end{abstract}

\section{Introduction}

\begin{figure*}[!t]
    \centering
    \includegraphics[width=\textwidth]{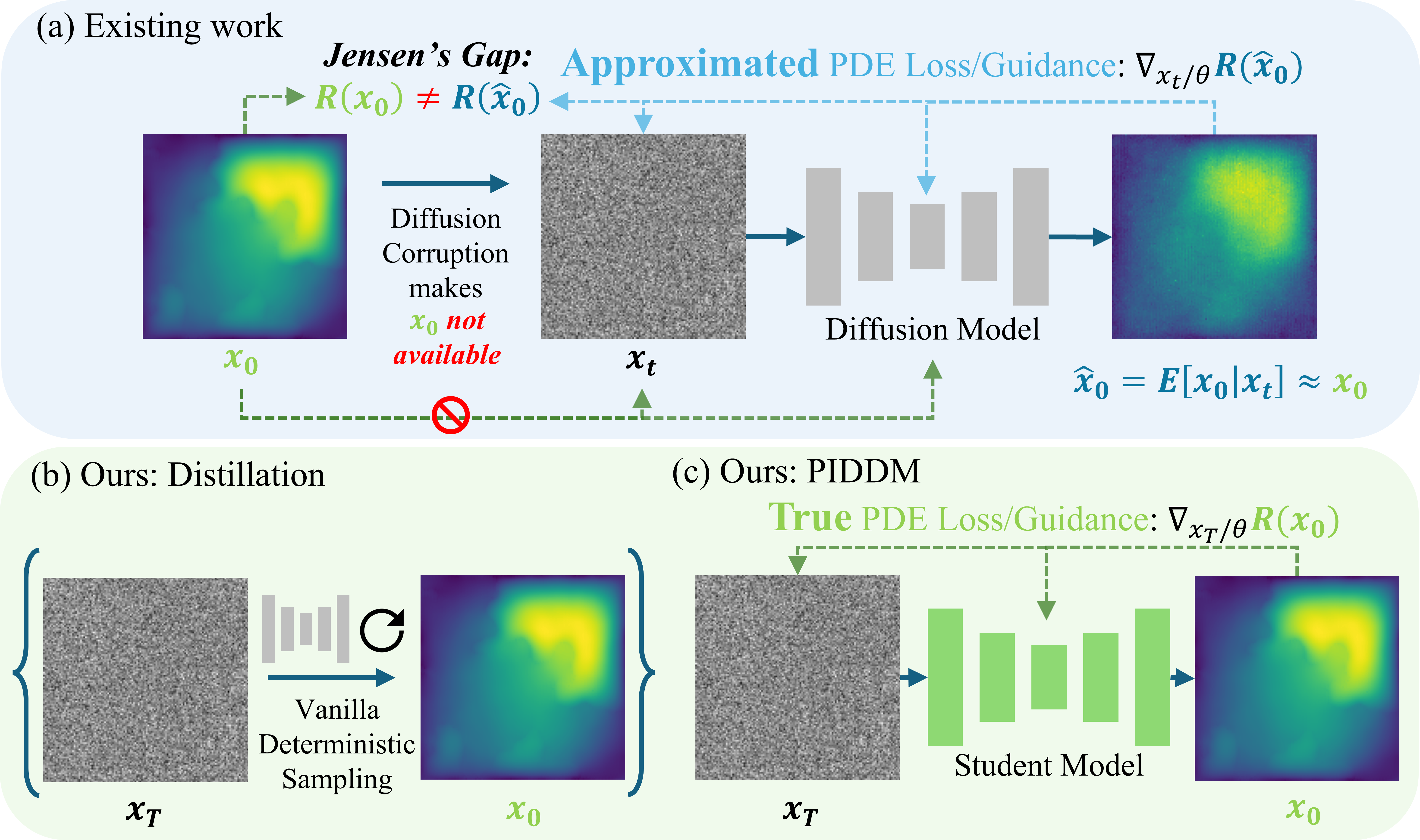}

    \caption{
\textbf{Illustration of physics-constrained diffusion generation and our proposed framework.}
\textbf{(a)} Existing methods \cite{huang2024diffusionpde, cheng2025eci, bastek2025pdim, jacobsen2024cocogen} impose PDE losses or guidance on the posterior mean $\mathbb{E}[x_0|x_t]$ in diffusion training and sampling, introducing the Jensen’s Gap.
\textbf{(b)} We train and sample a diffusion model using vanilla methods to generate a paired noise-data dataset for distillation.
\textbf{(c)} Our proposed framework distills the teacher diffusion model and directly enforces physical constraints on the final generated sample $x_0$, avoiding posterior-mean surrogate constraints in the student objective.}
    \label{fig:method}
    \vspace{-0.2cm}
\end{figure*}

Solving partial differential equations (PDEs) underpins innumerable applications in physics, biology, and engineering, spanning fluid flow \cite{davidson2015turbulence}, heat transfer \cite{incropera2011heat}, elasticity \cite{timoshenko1970elasticity}, electromagnetism \cite{jackson1998classical}, and chemical diffusion \cite{crank1975diffusion}. Classical discretization schemes such as finite-difference \cite{smith1985fem} and finite-element methods \cite{leveque2007fem} provide reliable solutions, but their computational cost grows sharply with mesh resolution, dimensionality, and parameter sweeps, limiting their practicality for large-scale or real-time simulations \cite{hughes2003finite}. This bottleneck has fueled a surge of learning-based solvers that approximate or accelerate PDE solutions, from early physics-informed neural networks (PINNs) \cite{raissi2019pinn} to modern operator-learning frameworks such as DeepONet \cite{lu2019deeponet}, Fourier Neural Operators \cite{li2020fno} and Physics-Informed Neural Operator \cite{li2024pino}, offering faster inference, uncertainty quantification, and seamless integration into inverse or data-driven tasks.


Among these learning-based solvers, diffusion models \cite{ho2020ddpm, song2020score} provide a promising framework for generative modeling of physical systems. For PDEs, a diffusion model can learn the joint distribution of solution and coefficient fields, $\vx_0=(\vu,\va)$, from data, where $\va$ denotes input parameters that satisfy the boundary operator $\mathcal{B}$ (for example, material properties or initial conditions) and $\vu$ is the corresponding solution that satisfies the PDE operator $\mathcal{F}$. After training, the model can sample $(\vu,\va)$ from this learned distribution, enabling forward simulation (sample $\vu$ given $\va$), inverse recovery (sample $\va$ given $\vu$), and conditional reconstruction (complete missing components of $\vu$ or $\va$) \emph{within a single framework, which prior non-diffusion approaches \cite{li2020fno, lu2019deeponet, raissi2019pinn} do not provide.} However, while diffusion models perform well under soft, high-level constraints \cite{rombach2022ldm, esser2024sd3, ho2022cfg, chung2022dps, ben2024dflow}, PDE applications often require strict, low-level constraints dictated by $\mathcal{F}$ and $\mathcal{B}$.



Enforcing such PDE constraints within diffusion models is nontrivial. A core difficulty is that, at an individual noise level $t$, diffusion models operate on noisy variables $\vx_t$ rather than the clean physical field $\vx_0$, where constraints such as $\mathcal{F}[\vx_0] = 0$ are defined. To address this, one option is to reconstruct $\vx_0$ by running the full deterministic sampling trajectory, but this is computationally expensive since it requires many forward passes, and enforcing constraints through backpropagation often causes gradient issues \cite{bastek2025pdim}. A more common alternative is to approximate $\vx_0$ with the posterior mean $\mathbb{E}[\vx_0 | \vx_t]$, which can be efficiently computed via Tweedie’s formula \cite{bastek2025pdim, huang2024diffusionpde, cheng2025eci, jacobsen2024cocogen, ddrm, ddnm, diffpir, daps, LGD, DPG, scg} (see the right panel of Fig.~\ref{fig:method}(a)). However, this introduces a mismatch: enforcing constraints on the posterior mean, $\mathcal{F}[\mathbb{E}[\vx_0 | \vx_t]]$, is not equivalent to enforcing the expected constraint, $\mathbb{E}[\mathcal{F}[\vx_0] | \vx_t]$, due to Jensen's inequality. This issue, known as the \textit{Jensen's Gap} in inverse-problem diffusion guidance \cite{chung2022dps} and later studied in PDE-constrained diffusion models \cite{bastek2025pdim, huang2024diffusionpde}, can lead to degraded physical fidelity.

\noindent\textbf{Contributions.}
We propose an effective framework that enforces PDE constraints in diffusion models via post-hoc distillation, enabling reliable and efficient generation under physical laws. As shown in Fig.~\ref{fig:method}(c), our method sidesteps the limitations of existing constraint-guided diffusion-based approaches by decoupling physics enforcement from the diffusion trajectory. Our main contributions are:

\begin{itemize}[leftmargin=*, nosep]
\item \textbf{Empirical confirmation of Jensen’s Gap:} We provide an empirical demonstration and quantitative analysis of the \textit{Jensen’s Gap} in PDE-constrained diffusion, a discrepancy that arises when PDE constraints are imposed on the posterior mean $\mathbb{E}[\vx_0 | \vx_t]$, rather than the final clean sample $\vx_0$.

\item \textbf{Final-sample physics enforcement:} PIDDM changes where and when physics is enforced: instead of guiding iterative denoising through posterior-mean surrogates, it imposes PDE supervision on the final output of a post-hoc distilled student. This removes the posterior-mean versus final-sample mismatch from the student objective, while our distributional fidelity claims remain empirical.

\item \textbf{Versatile and efficient inference:} The distilled student model preserves the full generative capabilities of the teacher, supporting physical simulation, reconstruction, and unified forward and inverse PDE solving within \emph{a single model}, while enabling one-step generation for fast inference. Experiments across diverse PDEs show that PIDDM surpasses posterior-mean-based methods \cite{huang2024diffusionpde, cheng2025eci, bastek2025pdim, jacobsen2024cocogen} in both generation quality and constraint satisfaction.
\end{itemize}

\begin{figure*}[!t]
    \begin{subfigure}{0.24\textwidth}
    \centering
    \includegraphics[width=\textwidth]{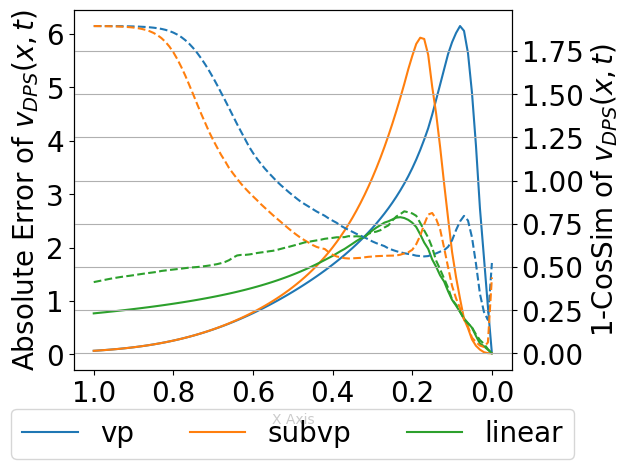}
    \vspace{-2em}\caption{}\label{fig:jensen_gap_mog:dps_analytical}
    \end{subfigure}\hfill
    \begin{subfigure}{0.24\textwidth}
    \centering
    \includegraphics[width=\textwidth]{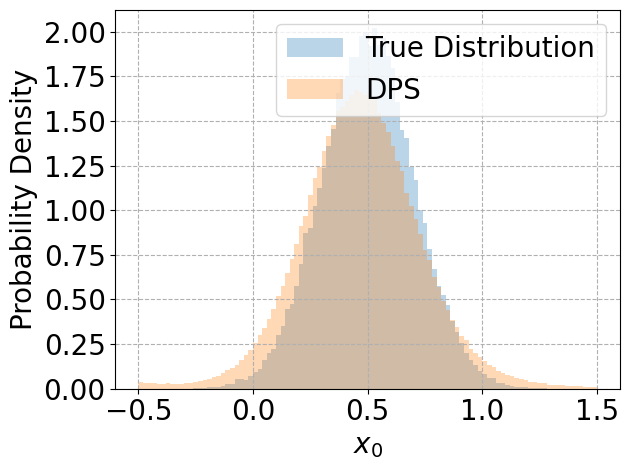}
    \vspace{-2em}
    \caption{}\label{fig:jensen_gap_mog:dps_x_0}
    \end{subfigure}\hfill
    \begin{subfigure}{0.23\textwidth}
    \centering
    \includegraphics[width=\textwidth]{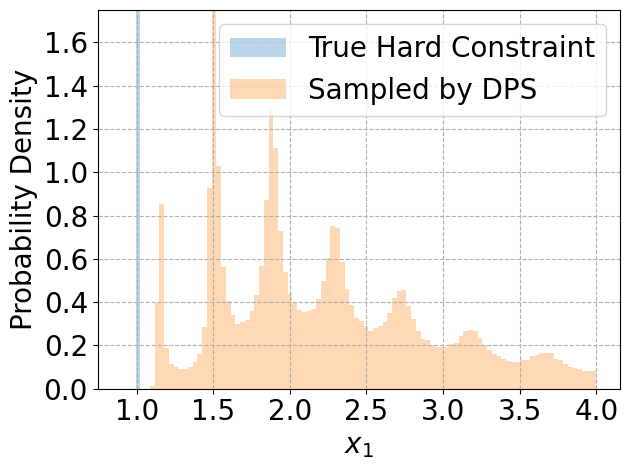}
    \vspace{-2em}
    \caption{}\label{fig:jensen_gap_mog:dps_x_1}
    \end{subfigure}\hfill
    \begin{subfigure}{0.24\textwidth}
    \centering
    \includegraphics[width=\textwidth]{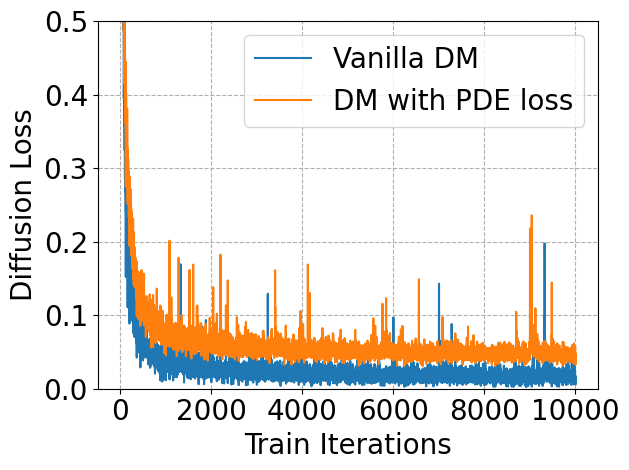}
    \vspace{-2em}
    \caption{}\label{fig:jensen_gap_mog:loss}
    \end{subfigure}\hfill
    \vskip -.05in
    \caption{
\textbf{Empirical illustration of the Jensen’s Gap in physics-constrained diffusion models.}
\textbf{(a)} Absolute velocity error and angular discrepancy ($1 - \cos(\theta)$) between Diffusion Posterior Sampling (DPS) and the ground-truth conditional ODE velocity on the MoG dataset. \textbf{(b)} and \textbf{(c)} Histograms comparing the first (unconstrained) and second (hard-constrained) dimensions of DPS-sampled MoG data against the ground-truth MoG. \textbf{(d)} Training-time manifestation: diffusion loss comparison on a Stokes problem dataset.
}\label{fig:jensen_gap_mog}
\end{figure*}

\section{Problem Setup: Jensen's Gap in Diffusion Models with PDE Constraints}\label{sec:Jensengap}

In scientific machine learning, there exist many \textit{hard} and \textit{low-level} constraints that are mathematically strict and non-negotiable \cite{leveque1992numerical, hansen2023learning, mouli2024using, saad2022guiding}. In this section, we discuss how existing works impose these constraints in diffusion-generated data, and the Jensen's Gap introduced by posterior-mean guidance in inverse problems \cite{chung2022dps} and inherited by PDE-constrained diffusion methods \cite{bastek2025pdim, huang2024diffusionpde}.

\subsection{Preliminaries on Physical Constraints}

Physical constraints are typically expressed as a \textit{partial differential equation (PDE)} $\mathcal{F}$ defined over a solution domain $\Omega \subset \mathbb{R}^d$, together with a \textit{boundary condition} operator $\mathcal{B}$ defined on the coefficient domain $\Omega^\prime$:
\begin{equation}
\mathcal{F}[\mathbf{u}(\boldsymbol{\xi})] = 0 \text{ for } \boldsymbol{\xi} \in \Omega, \quad \mathcal{B}[\mathbf{a}(\boldsymbol{\xi^\prime})] = 0 \text{ for } \boldsymbol{\xi^\prime} \in \Omega^\prime.
\end{equation}
In practice, the domains $\Omega$ and $\Omega^\prime$ are discretized into a uniform grid, typically of size $H \times W$, and the fields $\vu$ and $\va$ are evaluated at those grid points to produce the observed data $\vx_0=(\vu,\va)$, where diffusion models are trained to learn the joint distribution $p(\vx_0) = p((\vu, \va))$. While PINNs \cite{raissi2019pinn} model the mapping $\boldsymbol{\xi}\,,\boldsymbol{\xi^\prime} \mapsto (\mathbf{u}(\boldsymbol{\xi}),\mathbf{a}(\boldsymbol{\xi^\prime}))$ with differentiable neural networks to enable automatic differentiation \cite{paszke2019pytorch, abadi2016tensorflow}, grid-based approaches commonly approximate the differential operators in $\mathcal{F}$ via finite difference methods \cite{smith1985fem, leveque2007fem}. To quantify the extent to which a generated sample $\vx_0$ violates the physical constraints, the \textit{physics residual error} is often defined by:
\begin{equation}
\mathcal{R}(\vx_0)= \mathcal{R}((\vu, \va)):=
\big[
\mathcal{F}[\vu],
\mathcal{B}[\va]\big]^\top
\end{equation}
Here, $\mathcal{R}(\vx)$ measures the discrepancy between the sample $\vx$ and the expected PDE $\mathcal{F}$ and boundary conditions $\mathcal{B}$. The \textit{physics residual loss} is often defined by the squared norm of this physics residual error, i.e., $\Vert\mathcal{R}(\vx)\Vert^2$.

\subsection{Imposing PDE Constraints in Diffusion Models}
The physical constraints $\mathcal{R}$ are often defined on the clean field $\vx_0$, while during training or sampling of the diffusion model, the model only observes the noisy state $\vx_t$. Therefore, direct optimization or controlled generation based on the physical residual loss $\mathcal{R}(\vx_0)$ is generally impractical. A practical workaround is to evaluate the constraint on an estimate of $\vx_0$ from $\vx_t$, and a common choice is to use the estimated \emph{posterior mean}: $\mathbb{E}[\vx_0\mid\vx_t]$ based on the score network in a diffusion model \citep{bastek2025pdim,huang2024diffusionpde}. As a simplified example, consider the forward process defined as $\vx_t = \vx_0 + \sigma_t \boldsymbol{\epsilon}$, where $\sigma_t$ denotes the noise level at time $t$ and $\boldsymbol{\epsilon} \sim \mathcal{N}(0, \mathbf{I})$ is standard Gaussian noise. Then, the posterior mean can be efficiently estimated via Tweedie's formula:
\begin{align*}
\hat{\vx}_\theta(\vx_t, t) & := \vx_t + \sigma_t^2 s_\theta(\vx_t, t) \\
& \approx \vx_t + \sigma_t^2\nabla \log p(\vx_t) \approx \mathbb{E}[\vx_0 \mid \vx_t],
\end{align*}
where $s_\theta$ is a learned score function approximating the gradient of the log-density (see Appendix~\ref{appendix:devi-velocity} for the derivation for the general diffusion process).
Leveraging this approximation, several existing works incorporate PDE constraints by evaluating the PDE residual operator $\mathcal{R}$ on $\hat{\vx}_\theta(\vx_t, t)$. For instance, \textbf{PIDM}~\cite{bastek2025pdim} integrates PDE constraints into a diffusion model \textit{at training time} by augmenting the standard diffusion objective with an additional PDE residual loss $\mathcal R(\hat{\vx}_\theta(\vx_t, t))$. Similarly, \textit{at inference time}, \textbf{DiffusionPDE} \cite{huang2024diffusionpde} and \textbf{CoCoGen} \cite{jacobsen2024cocogen} employ diffusion posterior sampling (DPS)
 \cite{chung2022dps}, guiding each intermediate sample $\vx_t$ using the gradient $\nabla_{\vx} \mathcal R(\hat{\vx}_\theta(\vx_t, t))$. On the other hand, \textbf{ECI-sampling}~\cite{cheng2025eci} directly applies hard constraints to the posterior mean at each DDIM step using a correction operator (more detailed discussion on their implementations can be found in Appendix~\ref{appendix:baseline}). Beyond PDE applications, constrained diffusion models for image inverse problems also rely on posterior-mean approximations for true posterior estimation. Representative examples include DDRM~\cite{ddrm}, DDNM~\cite{ddnm}, LGD~\cite{LGD}, DPG~\cite{DPG}, SCG~\cite{scg}, DCDPM~\cite{dcdpm}, mid-point guidance~\cite{midpoint}, DiffPIR~\cite{diffpir}, and DAPS~\cite{daps} (see Appendix~\ref{Appendix:related-work:constrained-generation} for details). While these pioneering methods have been demonstrated to be effective in enforcing PDE constraints within diffusion models, they still suffer a theoretical inconsistency: PDE constraints are enforced on the posterior mean approximation $\mathbb{E}[\vx_0|\vx_t]$, which is not equivalent to the constraints on the true generated data $\vx_0$ due to Jensen's inequality:
\begin{equation}
\mathcal{R}(\mathbb{E}[\vx_0|\vx_t]) \neq \mathbb{E}[\mathcal{R}(\vx_0)|\vx_t].
\end{equation}
This discrepancy is commonly referred to as the \textit{Jensen's Gap} \cite{chung2022dps, bastek2025pdim, huang2024diffusionpde, gao2017jensen}. To mitigate this issue, PIDM and DiffusionPDE heuristically down-weight PDE constraints at early denoising steps (large $t$) in training and sampling, respectively, where the gap is pronounced, and emphasize them near $t \to 0$, where the posterior mean approximation improves. ECI-sampling introduces stochastic resampling steps \cite{wang2025resample} to project intermediate samples back toward their correct distribution. Many other methods for image inverse problems also provide partial improvements to reduce approximation error \cite{LGD, DPG, scg, dcdpm, midpoint} (Appendix~\ref{Appendix:related-work:constrained-generation}). These methods address aspects of the approximation error, but still rely on posterior-mean surrogates during the diffusion trajectory.


\subsection{Demonstration of the Jensen’s Gap}\label{sec:jensengap-demo}

To better illustrate this posterior-mean mismatch and its negative effect in PDE-constrained generation, we conduct experiments on two synthetic datasets: a Mixture-of-Gaussians (MoG) dataset and a Stokes Problem dataset.\\
\noindent\textbf{Sampling-time Jensen’s Gap.}
We demonstrate the sampling-time Jensen’s Gap using the Mixture-of-Gaussians (MoG) dataset, where the score function is analytically tractable, allowing us to isolate the effect of the diffusion process without interference from training error. The MoG is constructed in 2D: the first dimension follows a bimodal Gaussian distribution, while the second dimension encodes a discrete latent variable that serves as a hard constraint. Concretely, the joint distribution is defined as a mixture of two Gaussians, each supported on a distinct horizontal line:
\begin{align*}
p(x_0) = &\ 0.5 \cdot \, \mathcal{N}(x_1; -1, \sigma^2) \cdot \delta(x_2 + 1) + \\
&\ 0.5 \cdot \, \mathcal{N}(x_1; +1, \sigma^2) \cdot \delta(x_2 - 1),
\end{align*}
where $\delta(\cdot)$ denotes the Dirac delta function and $\sigma=0.2$. To examine the impact of Jensen’s Gap during sampling, we compare Diffusion Posterior Sampling (DPS) \cite{chung2022dps} which uses a latent code to guide the generation, with the ground-truth conditional ODE trajectory derived analytically.
We evaluate three representative diffusion processes: Variance-Preserving (VP) \cite{ho2020ddpm}, Sub-VP \cite{song2020score}, and Linear~\cite{liu2023flow}, and compare their velocity fields during inference to characterize Jensen's Gap.
To quantify amplitude errors, we compute the mean absolute error (MAE) and angular error between the DPS-predicted velocity field $v_{\text{DPS}}(x, t)$ and the ground-truth velocity $v_{\text{GT}}(x, t)$:
We observe that both DPS errors are significantly elevated at intermediate timesteps and only diminish as $t\to0$, as shown in Fig.~\ref{fig:jensen_gap_mog:dps_analytical}. Although DPS achieves accurate sampling in the unconstrained dimension (Fig.~\ref{fig:jensen_gap_mog:dps_x_0}), it fails to respect the hard constraint in the constrained dimension (Fig.~\ref{fig:jensen_gap_mog:dps_x_1}). This experiment illustrates the mismatch induced by posterior-mean guidance; it does not claim that single posterior samples along the trajectory resolve the gap. \\
\noindent\textbf{Training-time Jensen’s Gap.}
We examine the Jensen’s Gap during training using the synthetic Stokes dataset with target distribution $p_{\text{Stokes}}$. The diffusion model $\vv_\theta$ adopts a Fourier Neural Operator (FNO) \cite{li2020fno} architecture and follows a standard linear noise schedule \cite{lipman2023flow, liu2023rectifiedflow, liu2023instaflow}. Dataset and training details are given in Appendices~\ref{appendix:datasets} and~\ref{appendix:setup}. We take PIDM~\cite{bastek2025pdim} as a representative method, which augments the diffusion loss with a PDE residual term $\mathcal R(\hat\vx_\theta(\vx_t,t))$, and compare its performance with standard diffusion training. To assess generative quality, we monitor the diffusion loss, which theoretically corresponds to the evidence lower bound (ELBO) \cite{ho2020ddpm, kingma2013vae}.
The comparison results are shown in Fig.~\ref{fig:jensen_gap_mog:loss}, revealing a significant increase in diffusion loss when the PDE residual loss is incorporated. This suggests that the PDE residual loss does not improve modeling of the PDE-consistent data distribution. This observation also corroborates findings from PIDM \cite{bastek2025pdim}, which identified that residual supervision on the posterior mean can create ``a conflicting objective between the data and residual loss'', where the data loss represents the original diffusion training objective. These results provide further evidence of the posterior-mean mismatch in training, as enforcing constraints on $\mathbb{E}[\vx_0 | \vx_t]$ may interfere with maximizing the likelihood of the true data distribution.


\section{Method: Physics-Informed Distillation of Diffusion Models}
\begin{algorithm}[t]
\caption{PIDDM Training: Physics-Informed Distillation.}\label{alg:piddm_training}
\begin{algorithmic}[1]
\Require Teacher Model $\vv_{\theta}(\vx,t)$, Student Model $d_{\boldsymbol{\theta}'}$, Batch Size $B$, Terminal Time $T{=}1$, Steps $N_s$, Step Size $\mathrm{d}t{=}1/N_s$, Physics Residual Error $\mathcal R$, Loss Weight $\lambda_{\text{train}}$, Learning Rate $\eta_{\text{train}}$
\Repeat
  \State Sample $\boldsymbol{\epsilon}_{1:B}\stackrel{\text{i.i.d.}}{\sim}\cN(\mathbf 0,\mathbf I)$;\quad $\vx_T\leftarrow\boldsymbol{\epsilon}_{1:B}$
  \For{$t=T{-}\mathrm{d}t,\dots,0$}
    \State $\vx_t\leftarrow\vx_{t{+}\mathrm{d}t}-\vv_{\theta}(\vx_{t{+}\mathrm{d}t},t{+}\mathrm{d}t)\,\mathrm{d}t$ \Comment{\textbf{Sampling Phase}}
  \EndFor
  \State $\vx_{\text{pred}}\leftarrow d_{\boldsymbol{\theta}'}(\boldsymbol{\epsilon}_{1:B})$
  \State $\mathcal L\leftarrow\tfrac1B(\bigl\|\vx_{\text{pred}}-\vx_0\bigr\|^2+\lambda_{\text{train}}\bigl\|\mathcal R(\vx_{\text{pred}})\bigr\|^2)$ \Comment{\textbf{Distillation Phase}}
  \State $\theta'\leftarrow\theta'-\eta_{\text{train}}\nabla_{\theta'}\mathcal L$
\Until{Converged}
\end{algorithmic}
\end{algorithm}

In Section~\ref{sec:Jensengap}, we have demonstrated the posterior-mean mismatch that appears when incorporating physical constraints into diffusion training and sampling, as observed in prior works. To address this issue, we propose a distillation-based framework that sidesteps this mismatch in the student objective. Specifically, instead of enforcing constraints on the posterior mean during the diffusion process, we apply physical constraints directly to the final generated samples in a post-hoc distillation stage.

\subsection{Diffusion Training}
To decouple physical constraint enforcement from the diffusion process itself, we first conduct \textit{standard} diffusion model training using its original denoising objective, without adding any constraint-based loss. To obtain smoother sampling trajectories that benefit later noise-data distillation \cite{liu2023rectifiedflow, liu2023instaflow}, we adopt a linear diffusion process and apply the $\vv$-prediction parameterization \cite{liu2023rectifiedflow, lipman2023flow, liu2023instaflow, cheng2025eci, esser2024sd3}, which is commonly referred to as a flow model. Specifically, the training objective is
\begin{equation*}
    \mathcal{L}(\bm{\theta})
    = \mathbb{E}_{\substack{t\sim U(0, 1),\, \vx_0\sim p\\
    \bm{\epsilon} \sim \mathcal{N}(\mathbf{0}, \bm{I})}}
    \left[\left\| \vv_{\boldsymbol{\theta}}(\vx_t, t) - (\boldsymbol{\epsilon}-\vx_0) \right\|^2\right],
\end{equation*}
where $\vx_t = (1 - t) \vx_0 + t \boldsymbol{\epsilon}$, $p(\vx_0)$ is the distribution of joint data containing both solution and coefficient fields $\vx=(\vu, \va)$, $\boldsymbol{\epsilon}$ is sampled from a standard Gaussian distribution, and $\vv_{\boldsymbol{\theta}}$ is the neural network used as the diffusion model. This formulation allows the model to learn to reverse the diffusion process without entangling it with physical supervision, thereby preserving generative fidelity.

\subsection{Imposing PDE Constraints in Distillation}
\begin{algorithm}[t]
\caption{PIDDM Inference: Physics Data Simulation}
\label{alg:piddm_inference_generation}
\begin{algorithmic}[1]
\State \textbf{Input} Student Model $d_{\boldsymbol{\theta}'}$, Physics Residual Error $\mathcal{R}$, Refinement Step Number $N_f$, Refinement Step Size $\eta_{\text{ref}}$, Latent Noise $\boldsymbol{\epsilon} \sim \mathcal{N}(\mathbf{0},\mathbf{I})$.

\For{$i = 1, \dots, N_f$}
    \State $\boldsymbol{\epsilon} \gets \boldsymbol{\epsilon} - \eta_{\text{ref}}\nabla_{\boldsymbol{\epsilon}}\Vert\mathcal{R}(d_{\boldsymbol{\theta}'}(\boldsymbol{\epsilon}))\Vert^2$
\EndFor
\State \textbf{Output} $d_{\boldsymbol{\theta}'}(\boldsymbol{\epsilon})$
\end{algorithmic}
\end{algorithm}


\begin{algorithm}[!t]
\caption{PIDDM Inference for Forward/Inverse/Reconstruction}
\label{alg:piddm_downstream}
\begin{algorithmic}[1]
\State \textbf{Input} Student Model $d_{\boldsymbol{\theta}'}$, Physics Residual Error $\mathcal{R}$, Optimization Iterations $N_o$, Step Size $\eta_{\text{infer}}$, Observation $\vx'$, Observation Mask $M$, Loss Weight $\lambda_{\text{infer}}$, Latent Noise $\boldsymbol{\epsilon} \sim \mathcal{N}(\mathbf{0},\mathbf{I})$.
\For{$i = 1, \dots, N_o$}
    \State $\vx_{\text{mix}} \gets \vx' \odot M + d_{\boldsymbol{\theta}'}(\boldsymbol{\epsilon}) \odot (1 - M)$
    \State $\boldsymbol{\epsilon} \gets \boldsymbol{\epsilon} - \eta_{\text{infer}} \nabla_{\boldsymbol{\epsilon}}[\left\| (d_{\boldsymbol{\theta}'}(\boldsymbol{\epsilon}) - \vx') \odot M \right\|^2 + \lambda_{\text{infer}} \left\| \mathcal{R}(\vx_{\text{mix}}) \right\|^2]$
\EndFor
\State $\vx \gets \vx' \odot M + d_{\boldsymbol{\theta}'}(\boldsymbol{\epsilon}) \odot (1 - M)$
\State \textbf{Output} $\vx$
\end{algorithmic}
\end{algorithm}

\begin{table*}[t]
\vspace{-.2cm}
\centering
\small
\setlength{\tabcolsep}{3pt}
\renewcommand{\arraystretch}{1.2}
\caption{Generative metrics on various PDE problems. PDE error is the MSE of the evaluated physics residual error.
         The best results are in \textbf{bold} and the second-best results are \underline{underlined}.}
\label{tab:generative_evaluations}
\begin{tabular}{ll
                S  S  S  S  S  S  S}
\toprule
Dataset & Metric &
  \textbf{PIDDM-1} & \textbf{PIDDM-ref} & {ECI} & {DiffusionPDE} &
  {D-Flow} & {PIDM} & {Teacher}\\
\midrule
\multirow{5}{*}{Darcy}
  & MMSE ($\times 10^{-2}$) &
    {\underline{0.112}} & {\bfseries 0.037} & {0.153} & {0.419} & {0.129} & {0.515} & {0.108} \\
  & SMSE ($\times 10^{-2}$) &
    {\underline{0.082}} & {\bfseries 0.002} & {0.103} & {0.163} & {0.085} & {0.368} & {0.069} \\
  & PDE Error ($\times 10^{-4}$) &
    {\underline{0.226}} & {\bfseries 0.148} & {1.582} & {1.071} & {0.532} & {1.236} & {1.585} \\
  & FPD  &
    {\underline{0.754}} & {\bfseries 0.385} & {0.921} & {1.437} & {0.995} & {1.983} & {0.782} \\
  & NFE ($\times 10^{3}$) &
    {\bfseries 0.001} & {\underline{0.080}} & {0.500} & {0.100} & {5.000} & {0.100} & {0.100} \\
\midrule
\multirow{5}{*}{Poisson}
  & MMSE ($\times 10^{-2}$) &
    {\underline{0.162}} & {\bfseries 0.113} & {0.183} & {0.861} & {0.172} & {0.948} & {0.150} \\
  & SMSE ($\times 10^{-2}$) &
    {\underline{0.326}} & {\bfseries 0.274} & {0.291} & {0.483} & {0.475} & {0.701} & {0.353} \\
  & PDE Error ($\times 10^{-9}$) &
    {\underline{0.073}} & {\bfseries 0.050} & {2.420} & {1.270} & {0.831} & {1.593} & {2.443} \\
  & FPD  &
    {\underline{1.281}} & {\bfseries 0.659} & {1.532} & {1.835} & {1.677} & {2.358} & {1.342} \\
  & NFE ($\times 10^{3}$) &
    {\bfseries 0.001} & {\underline{0.080}} & {0.500} & {0.100} & {5.000} & {0.100} & {0.100} \\
\midrule
\multirow{5}{*}{Burgers}
  & MMSE ($\times 10^{-2}$) &
    {\underline{0.152}} & {\bfseries 0.012} & {0.294} & {0.064} & {0.305} & {0.948} & {0.264} \\
  & SMSE ($\times 10^{-2}$) &
    {\underline{0.133}} & {\bfseries 0.101} & {0.105} & {0.103} & {0.207} & {0.701} & {0.114} \\
  & PDE Error ($\times 10^{-3}$) &
    {\underline{0.466}} & {\bfseries 0.174} & {1.572} & {1.032} & {0.730} & {1.593} & {1.334} \\
  & FPD  &
    {\underline{0.129}} & {\bfseries 0.054} & {0.387} & {1.133} & {0.695} & {1.437} & {0.118} \\
  & NFE ($\times 10^{3}$) &
    {\bfseries 0.001} & {\underline{0.080}} & {0.500} & {0.100} & {5.000} & {0.100} & {0.100} \\
\bottomrule
\end{tabular}
\end{table*}

After training the teacher diffusion model using the standard denoising objective, we proceed to the distillation stage, where we transfer its knowledge to a student model designed for efficient one-step generation. Crucially, this post-hoc distillation stage is where we impose PDE constraints, moving physics enforcement away from posterior-mean surrogates used during diffusion training or sampling. This distillation process is guided by two complementary objectives: (1) learning to map a noise sample to the final generated output predicted by the teacher model, and (2) enforcing physical consistency on this output via PDE residual minimization. Concretely, we begin by sampling a noise input $\bm \epsilon \sim \mathcal{N}(\bm 0, \bm{I})$ and generating a target sample $x_0$ using the pre-trained teacher model via deterministic integration of the reverse-time ODE:
\begin{equation}
\vx_{t-\mathrm{d}t} = \vx_{t} - \vv_{\boldsymbol{\theta}}(\vx_{t}, t)\,\mathrm{d}t,
\end{equation}
which proceeds from $t=1$ to $t=0$ using a fixed step size $dt$. This yields a paired noise-data dataset $\mathcal{D}=\{\boldsymbol{\epsilon},\vx_0\}$ for distillation, as shown in Fig.~\ref{fig:method}(b). A student model $d_{\boldsymbol{\theta}'}(\boldsymbol{\epsilon})$ is then trained to predict $\vx_0$ in one step, as shown in Fig.~\ref{fig:method}(c). Meanwhile, to enforce physical consistency, we evaluate the physics residual error on the output $\vx = d_{\boldsymbol{\theta}'}(\bm \epsilon)$, i.e., $\left\|\mathcal{R}(\vx)\right\|^2$. The overall training objective is:
\begin{equation}\label{eq:distillation_loss}
\begin{aligned}
\mathcal{L}_{\text{total}} & = \mathcal{L}_{\text{sample}} + \lambda_{\text{train}} \mathcal{L}_{\text{PDE}}  \\
& =\mathbb{E}_{(\boldsymbol{\epsilon},\vx_0)\sim \mathcal{D}} \left[\left\| d_{\boldsymbol{\theta}'}(\boldsymbol{\epsilon}) - \vx_0 \right\|^2 + \lambda_{\text{train}}\left\| \mathcal{R}(\vx) \right\|^2\right],
\end{aligned}
\end{equation}
where $\lambda_{\text{train}}$ balances generative fidelity and physical constraint satisfaction. Unlike prior work \cite{bastek2025pdim, huang2024diffusionpde}, the residual is evaluated on final student outputs rather than posterior-mean surrogates $\mathbb{E}[\bm x_0 | \bm x_t]$, which makes the physical objective directly aligned with sample-level constraint satisfaction (see Table~\ref{tab:ablation_more_distillation}). Training is repeated until convergence (Algorithm~\ref{alg:piddm_training}).
This noise-to-sample distillation is often difficult to learn due to the high curvature of sampling trajectories, yielding noise–data pairs that are far apart in Euclidean space~\cite{liu2023instaflow}. To reduce curvature and improve learnability, we adopt linear-flow distillation~\cite{lipman2023flow, liu2023rectifiedflow}, and we further evaluate Distribution Matching Distillation (DMD)~\cite{yin2024dmd}, Rectified Flow, and Consistency Model \cite{song2023consistencymodels} to strengthen coupling and distribution alignment. These choices produce consistent gains (see Table~\ref{tab:ablation_more_distillation}).

To clarify the optimization mechanism, Appendix~\ref{app:pf-sampling} analyzes a low-rank Gaussian target, where the low-rank direction mimics a hard equality constraint in PDE problems. The main intuition is that diffusion training and posterior-mean guidance operate in score space along a noisy trajectory. Near the clean endpoint, the score component associated with the constrained direction becomes increasingly sharp, so small approximation errors in this component can persist as visible constraint violations in generated samples. PIDDM instead moves the physics penalty to data space after the endpoint sample is formed. The student therefore does not need to resolve the sharp score field at every noise level; it is trained to suppress the violating component of its final output directly. This explains why final-sample distillation can reduce sample-level PDE residuals, while our distributional fidelity claims remain empirical rather than an end-to-end theoretical guarantee.

\subsection{Downstream Tasks}\label{subsec:methods:downstream}

Our method naturally supports one-step generation of physically constrained data, jointly producing both coefficient and solution fields. Beyond this intrinsic functionality, it also retains the flexibility of the teacher diffusion model, enabling various downstream tasks such as forward and inverse problem solving, and reconstruction from partial observations. Compared to the teacher model, our method achieves these capabilities with improved computational efficiency and stronger physical alignment.

\noindent\textbf{Generative Modeling.}
We aim to sample physically consistent pairs $\vx_0=(\vu,\va)$ from a learned distribution that satisfies the governing PDE system. The student model supports this via efficient one-step generation: given $\boldsymbol{\epsilon}\sim\mathcal{N}(\mathbf{0},\bm{I})$, it outputs $\vx_0=d_{\boldsymbol{\theta}'}(\boldsymbol{\epsilon})$, approximating a valid solution–coefficient pair. We further provide an optional refinement stage based on constraint-driven optimization (Algorithm~\ref{alg:piddm_inference_generation}), which reduces the physics residual by updating $\boldsymbol{\epsilon}$ with gradient descent. This design is inspired by noise prompting methods~\cite{ben2024dflow, guo2024initno_goldennoise} that optimize the final sample with respect to the initial noise. However, in contrast to those prior works, which backpropagate through an entire sampling trajectory and incur high cost and unstable gradients, our refinement operates in a one-step setting. While optional, it offers additional control that is useful in scientific applications requiring strict physical consistency~\cite{leveque1992numerical, hansen2023learning, mouli2024using, saad2022guiding}.

\noindent\textbf{Forward/Inverse Problem and Reconstruction.}
PIDDM handles all downstream problems as conditional generation over the joint field \(\vx=(\vu,\va)\). Forward inference draws \(\vu\) from known \(\va\); inverse inference recovers \(\va\) from observed \(\vu\); reconstruction fills in missing entries of \((\vu,\va)\) given a partial observation \(\vx'\). We solve this via optimization-based inference on the latent variable \(\boldsymbol{\varepsilon}\), using the same student model \(d_{\boldsymbol{\theta}'}\) as in generation, as described in Algorithm~\ref{alg:piddm_downstream}. Let \(\vx = d_{\boldsymbol{\theta}'}(\boldsymbol{\varepsilon})\) denote the generated sample, and let \(M\) be a binary observation mask indicating the known entries in \(\vx'\) with respect to $\vx$. To ensure hard consistency with observed values (e.g., boundary conditions \(\mathcal{B}\)), we define a mixed sample by injecting observed entries into the generated output, following ECI-sampling~\cite{cheng2025eci}, and then update $\varepsilon$ using a combined objective:
\begin{equation}\label{eq:hard_constraint}
\begin{aligned}
\mathcal{L}_{\text{total}} &= \left\| \left(\vx - \vx'\right) \odot M \right\|^2
    + \lambda \left\| \mathcal{R}(\vx_{\text{mix}}) \right\|^2, \\
\vx_{\text{mix}} &= \vx' \odot M + \vx \odot (1 - M).
\end{aligned}
\end{equation}
Interestingly, we also find that applying this masking not only enhances hard constraints on $\mathcal{B}$, but also improves satisfaction of $\mathcal{F}$, as demonstrated in our ablation study in Table~\ref{tab:ablation_more_distillation}.
Classical inverse solvers~\cite{li2020fno,li2024pino,lu2019deeponet,raissi2019pinn} learn a deterministic map \(\vu\!\mapsto\!\va\) and therefore require full observations of \(\va\) to evaluate \(\mathcal{F}[\vu,\va]=0\), a condition rarely met in practice. DiffusionPDE~\cite{huang2024diffusionpde} relaxes this by sampling missing variables, but enforces physics on the posterior mean, i.e.\ \(\mathcal{F}\bigl[\mathbb{E}[\vx_0|\vx_t]\bigr]\), and thus suffers from the Jensen’s Gap. Our method avoids this inconsistency by imposing constraints directly on the final sample \(\mathcal{F}[\vx_0]\), yielding more reliable and physically consistent inverse solutions.

\section{Experiments}

\noindent\textbf{Experiment Setup.}
We consider three widely used PDE benchmarks in the main text: Darcy flow, Poisson equation, and Burgers' equation. These datasets are readily accessible from FNO \cite{li2020fno} and DiffusionPDE \cite{huang2024diffusionpde}. We also provide results on other benchmarks in Appendix~\ref{appendix:results_more_data}. We consider ECI \cite{cheng2025eci}, DiffusionPDE \cite{huang2024diffusionpde}, D-Flow \cite{cheng2025eci, ben2024dflow}, PIDM \cite{bastek2025pdim}, and vanilla teacher diffusion models as baseline methods, with detailed implementations in Appendix~\ref{appendix:baseline}. We follow ECI-sampling \cite{cheng2025eci} to use FNO as both the teacher diffusion model and the student distillation model. We provide the full specification of our experiment setup in Appendix~\ref{appendix:setup}.

To quantitatively evaluate generative performance, we report MMSE, SMSE, FPD, and PDE error following prior work \cite{cheng2025eci, kerrigan2023functional, bastek2025pdim, jacobsen2024cocogen}: MMSE measures the mean squared error of the sample mean; SMSE evaluates the error of the sample standard deviation; FPD evaluates the Frechet distance between hidden representations extracted by the pre-trained PDE foundation model; PDE error quantifies the violation of physical constraints using the physics residual error $\lvert \mathcal{R}(\vx)\rvert^2$. The number of function evaluations (NFE) reflects computational cost during inference. For downstream tasks, we further report MSE on solution fields, coefficient fields, or both, depending on the problem setting, reflecting the accuracy of PDE solving.

\subsection{Empirical Evaluations}
PIDDM samples the joint field \((\vu,\va)\), enabling forward (\(\vu|\va\)), inverse (\(\va|\vu\)), and reconstruction (partial \(\vu,\va\)) tasks (Sec.~\ref{subsec:methods:downstream}). DiffusionPDE~\cite{huang2024diffusionpde} reports only reconstruction MSE, while ECI-sampling~\cite{cheng2025eci} and PIDM~\cite{bastek2025pdim} cover at most one task, limited to either unconditional generation or forward solving. For a fair comparison, we evaluate all methods on all three tasks, providing a unified view of generative quality and physical fidelity.

\begin{table*}[t]
\centering
\setlength{\tabcolsep}{3pt}
\renewcommand{\arraystretch}{1.2}
\caption{Evaluation on various downstream tasks on Darcy datasets. PDE error is the MSE of the evaluated physics residual error. The units of MSE, PDE error, and NFE are $\times 10^{-1}$, $\times 10^{-4}$, and $\times 10^3$, respectively. The best results are in \textbf{bold}.}
\label{tab:downstream_evaluations}
\begin{tabular}{ll
                S  S  S  S  S}
\toprule
Task & Metric & \textbf{PIDDM} & {ECI} & {DiffusionPDE} & {D-Flow} & {PIDM}\\
\midrule
\multirow{3}{*}{Forward}
  & MSE  &
                     {\bfseries0.316} & {0.776} & {0.691} & {0.539} & {{0.380}} \\
  & PDE Error  &
                     {\bfseries0.145} & {1.573} & {1.576} & {{0.584}} & {1.248} \\
  & NFE  &
                     {\bfseries0.080} & {0.500} & {{0.100}} & {5.000} & {{0.100}} \\
\midrule
\multirow{3}{*}{Inverse}
  & MSE  &
                     {\bfseries 0.236} & {0.545} & {0.456} & {{0.428}} & {0.468} \\

  & PDE Error  &
                     {\bfseries0.126} & {1.505} & {1.402} & {{0.438}} & {1.113} \\
  & NFE  &
                     {\bfseries0.080} & {0.500} & {{0.100}} & {5.000} & {{0.100}} \\
\midrule
\multirow{4}{*}{Reconstruct}
  & Coef MSE  &
                     {\bfseries 0.128} & {0.395} & {0.240} & {{0.158}} & {0.179} \\
  & Sol MSE  &
                     {\bfseries 0.102} & {0.219} & {0.143} & {{0.125}} & {0.147} \\
  & PDE Error  &
                     {\bfseries 0.143} & {1.205} & {1.239} & {{0.605}} & {1.240} \\
  & NFE  &
                     {\bfseries0.080} & {0.500} & {{0.100}} & {5.000} & {{0.100}} \\
\bottomrule
\end{tabular}
\end{table*}

\begin{figure*}[t]
    \centering
    \begin{subfigure}{0.33\textwidth}
    \centering
    \includegraphics[width=\textwidth]{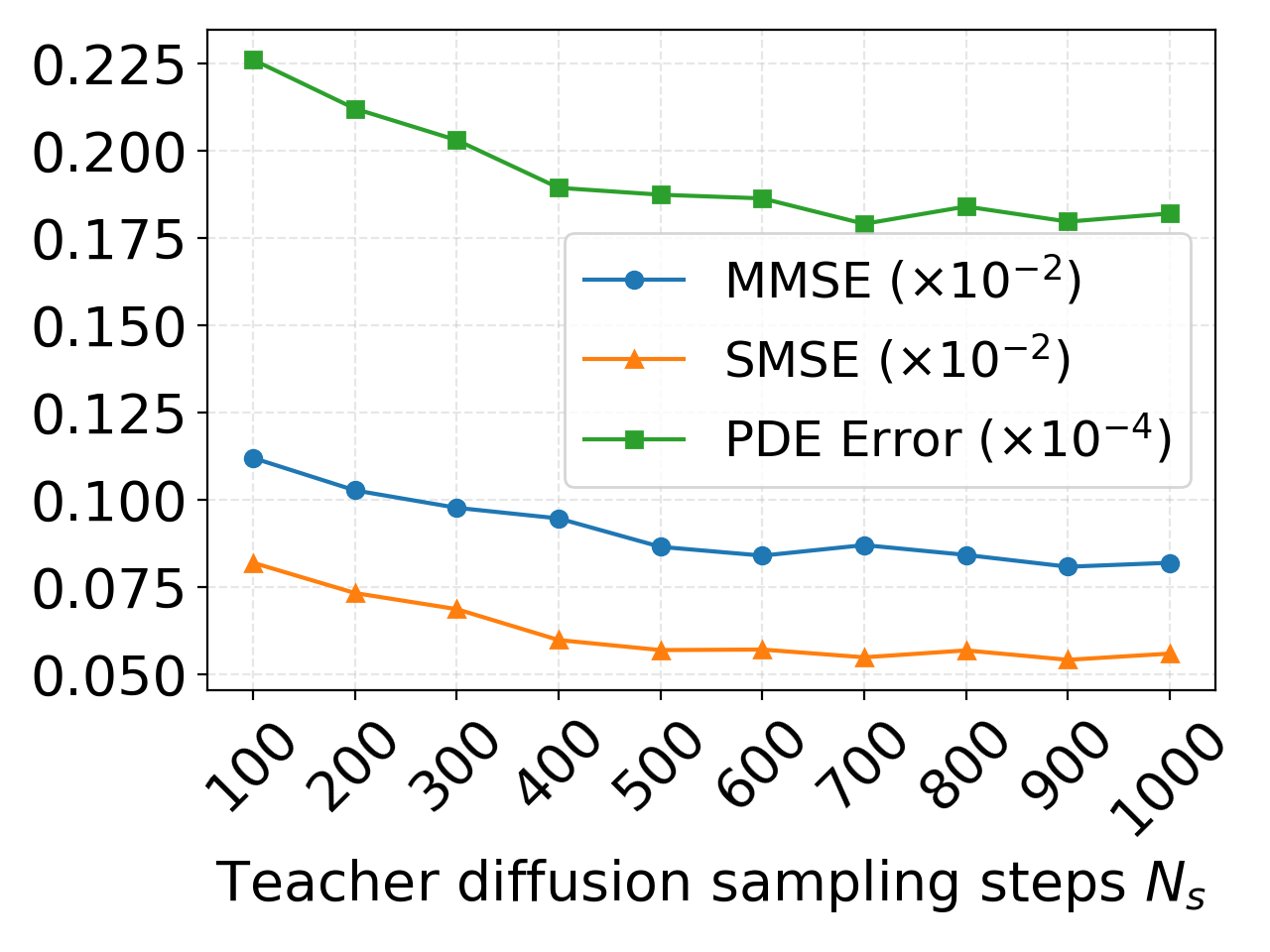}
    \vspace{-2em}
    \caption{}
    \end{subfigure}\hfill
    \begin{subfigure}{0.33\textwidth}
    \centering
    \includegraphics[width=\textwidth]{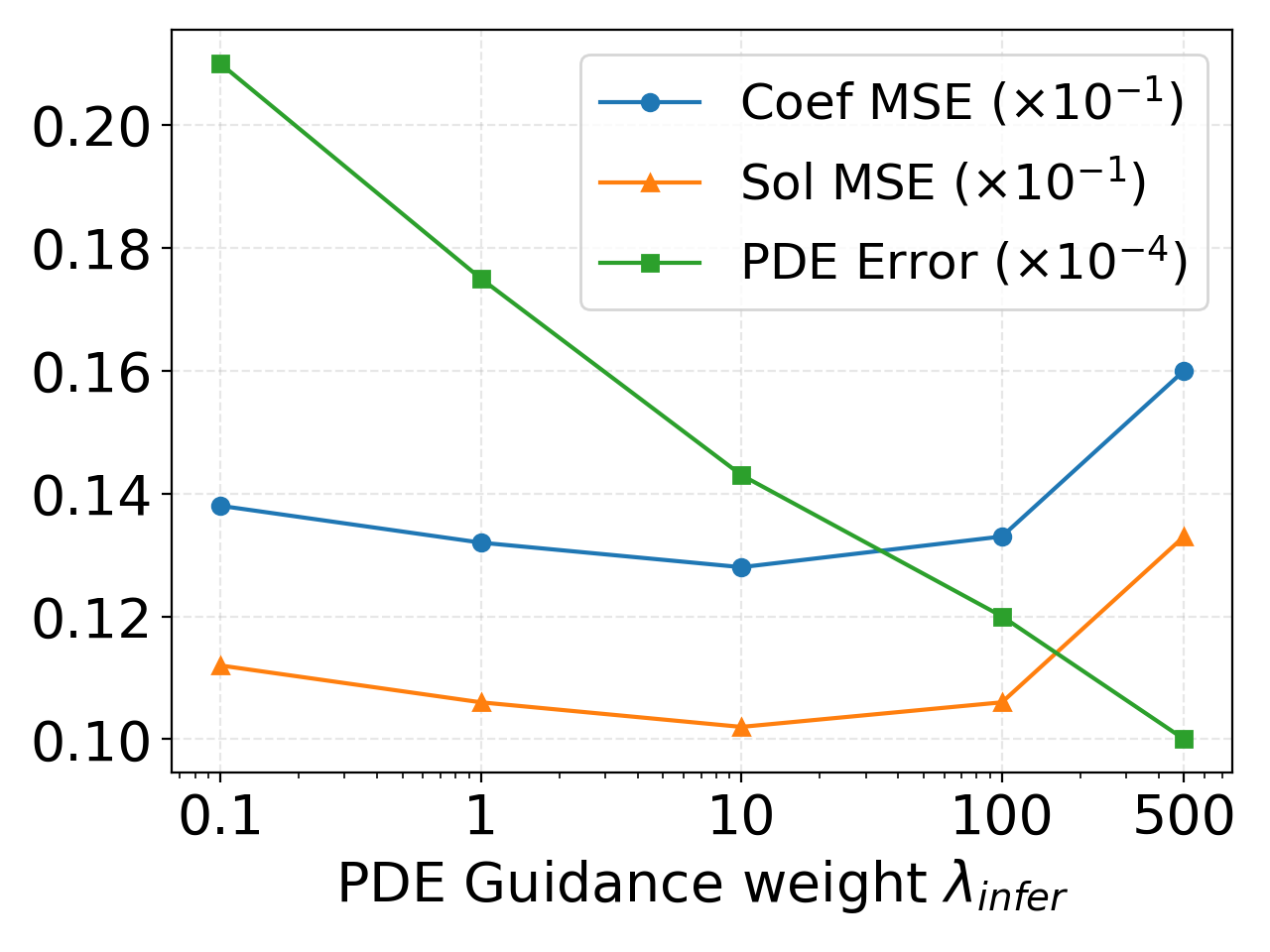}
    \vspace{-2em}
    \caption{}
    \end{subfigure}\hfill
    \begin{subfigure}{0.33\textwidth}
    \centering
    \includegraphics[width=\textwidth]{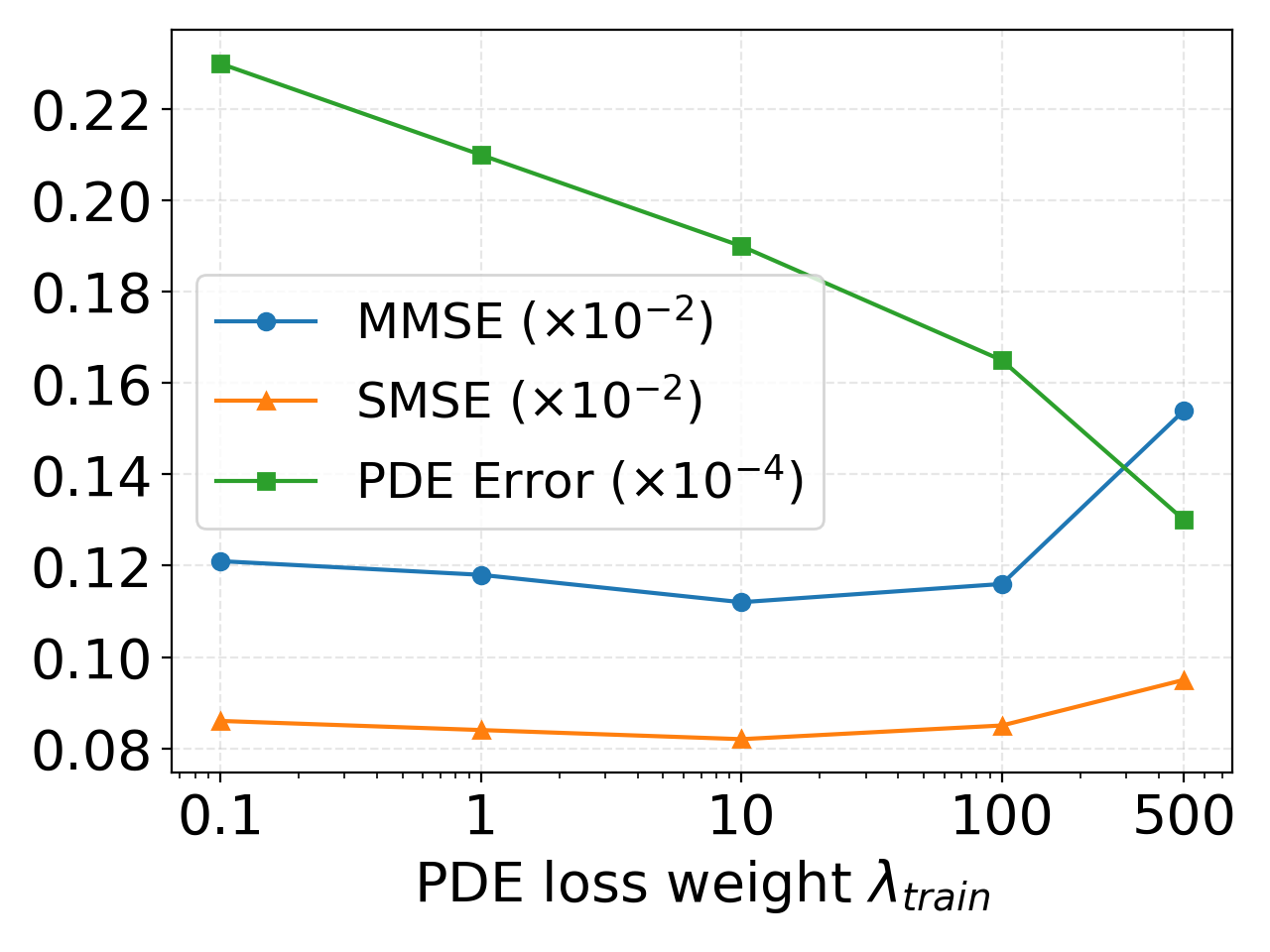}
    \vspace{-2em}
    \caption{}
    \end{subfigure}
    \vspace{-.2cm}
    \caption{Ablation studies on the effect of several factors on the performance of PIDDM on the Darcy dataset. Panels (a), (b), and (c) show the effects of $N_s$, $\lambda_{\text{train}}$, and $\lambda_{\text{infer}}$, respectively.}
    \label{fig:ablation_study}
\end{figure*}

\noindent\textbf{Generative Tasks.}
We first evaluate the generative performance of our method across three representative PDE systems: Darcy, Poisson, and Burgers' equation. As shown in Table~\ref{tab:generative_evaluations}, our one-step model (\textit{PIDDM-1}) achieves competitive MMSE and SMSE scores while maintaining extremely low computational cost (1 NFE). Notably, \emph{\textit{PIDDM-1} already surpasses prior methods that incorporate physical constraints during training or sampling}, such as PIDM, DiffusionPDE, and ECI-sampling, which impose constraints on posterior-mean surrogates and only exhibit marginal improvements over vanilla diffusion baselines. \emph{Our optional refinement stage (\textit{PIDDM-ref}) further reduces both statistical errors and physical PDE residuals, outperforming all baselines.} Meanwhile, ECI, which only enforces hard constraints on boundary conditions, achieves moderate improvements but remains less effective on field-level physical consistency. Although D-Flow enforces physical constraints through final-sample trajectory optimization, it requires thousands of NFEs and often suffers from gradient instability.

\begin{table*}[t]
\centering
\renewcommand{\arraystretch}{1.08}
\caption{Compute-matched Darcy generation. Best or tied-best entries are in \textbf{bold}.}
\label{tab:compute_matched}
\begin{tabular}{@{}lrrrrrrr@{}}
\toprule
Method & Time & NFE & BWD & PDE & FPD & MMSE & SMSE \\
\midrule
PIDDM-1 & \bfseries 0.004 & \bfseries 1 & \bfseries 0 & 0.226 & 0.754 & 0.112 & 0.082 \\
PIDDM-ref & 1.170 & 80 & 80 & \bfseries 0.148 & \bfseries 0.385 & \bfseries 0.037 & \bfseries 0.002 \\
DiffPDE & 1.250 & 80 & 80 & 1.129 & 1.452 & 0.431 & 0.173 \\
PIDM & 1.281 & 300 & \bfseries 0 & 1.274 & 2.018 & 0.522 & 0.374 \\
D-Flow & 19.302 & 2000 & 2000 & 0.885 & 1.252 & 0.254 & 0.209 \\
ECI & 1.294 & 300 & \bfseries 0 & 1.106 & 0.837 & 0.138 & 0.112 \\
\bottomrule
\end{tabular}
\par\vspace{1mm}
\begin{minipage}{\columnwidth}
\footnotesize Time is wall-clock seconds per sample; PDE, MMSE, and SMSE are scaled as in Table~\ref{tab:generative_evaluations}.
\end{minipage}
\end{table*}

\noindent\textbf{Compute-Matched Comparison.}
Table~\ref{tab:compute_matched} further compares inference cost and quality on Darcy generation under the same hardware. PIDDM-1 uses one forward evaluation and no backward pass, giving millisecond-level generation while improving PDE loss and distributional metrics over posterior-mean-guided baselines. With the same 80-step online budget as DiffusionPDE, PIDDM-ref uses final-sample refinement to obtain lower PDE loss, FPD, MMSE, and SMSE. D-Flow optimizes the final sample directly but requires 2000 forward and backward passes, making it substantially slower despite weaker statistical metrics.
Here, NFE denotes the number of forward function evaluations, and BWD denotes the number of inference-time backward passes.

\noindent\textbf{Forward/Inverse Solving and Reconstruction.}
We further demonstrate the versatility of our method in forward and inverse problem solving on the Darcy dataset. Since the original PIDM \cite{bastek2025pdim} implementation addresses only unconditional generation, we pair it with Diffusion Posterior Sampling (DPS) \cite{chung2022dps} to extend it to downstream tasks (forward, inverse, and reconstruction). Following the test protocol of D-Flow, we apply inference-time optimization over the initial noise to match given observations while satisfying physical laws. As shown in Table~\ref{tab:downstream_evaluations}, \emph{our method (\textbf{PIDDM}) achieves the best results across all metrics}, including MSE and PDE error, while being significantly more efficient than D-Flow, which requires 5000 function evaluations. Compared to ECI and DiffusionPDE, our method yields lower residuals and better predictive accuracy, reflecting its superior handling of physical and observational constraints jointly.

\begin{table*}[t]
\centering
\setlength{\tabcolsep}{3pt}
\renewcommand{\arraystretch}{1.2}
\caption{Evaluation on various downstream tasks on Darcy datasets under different PIDDM settings. PIDDM denotes the raw method, $+{\text{RF-1}}$ and $+{\text{RF-2}}$ denote one and two rounds of reflowing \cite{liu2023instaflow}, $+{\text{DMD}}$ denotes distribution matching distillation \cite{yin2024dmd}, and $+{\text{CM}}$ denotes the consistency model \cite{song2023consistencymodels}. {$-{\text{HC}}$} refers to the ablation without the hard-constraint replacement in Eq.~\ref{eq:hard_constraint} during PIDDM inference. {${\text{VP}}$} and {${\text{sub-VP}}$} refer to ablations on VP and sub-VP diffusion processes. PDE error is the MSE of the evaluated physics residual error. The best results are in \textbf{bold}.}
\label{tab:ablation_more_distillation}
\begin{tabular}{ll
                S  S  S  S  S  S  S  S}
\toprule
Task & Metric & {PIDDM} & {$+{\text{RF-1}}$} & {+{\text{RF-2}}} & {$+{\text{DMD}}$} & {$+{\text{CM}}$} & {$-{\text{HC}}$} & {${\text{VP}}$} & {${\text{sub-VP}}$} \\
\midrule
\multirow{2}{*}{Forward}
  & MSE ($\times 10^{-1}$) &
                    {0.316} & {0.278} & {\bfseries0.127} & {0.255} & {0.283} & {0.705} & {0.398} & {0.372}\\
  & PDE Error ($\times 10^{-4}$) &
                    {0.145} & {0.129} & {0.098} & {0.134} & {\bfseries0.083}& {0.354} & {0.154} & {0.157}\\
\midrule
\multirow{2}{*}{Inverse}
  & MSE ($\times 10^{-1}$) &
                    { 0.236} & {0.195} & {\bfseries0.136} & {0.188} & {0.182} & {0.503} & {0.284} & {0.271}\\
  & PDE Error ($\times 10^{-4}$) &
                    {0.115} & {0.126} & {\bfseries0.079} & {0.121} & {0.109} & {0.321} & {0.143} & {0.139}\\
\midrule
\multirow{3}{*}{Reconstruct}
  & Coef MSE ($\times 10^{-1}$) &
                    { 0.128} & {0.107} & {0.091} & {0.095} & {\bfseries 0.085} & {0.294} & {0.133} & {0.138}\\
  & Sol MSE ($\times 10^{-1}$) &
                    { 0.102} & {0.084} & {\bfseries0.063} & {0.073} & {0.072} &  {0.239} & {0.127} & {0.119}\\
  & PDE Error ($\times 10^{-4}$) &
                    { 0.143} & {0.118} & {0.085} & {0.104} & {\bfseries0.083} & {0.464} & {0.159} & {0.158} \\
\bottomrule
\end{tabular}
\end{table*}

\subsection{Ablation Studies}\label{sec:ablation}

To better understand the effect of key design choices in PIDDM, we perform ablations on five factors: teacher sampling steps \(N_s\); distillation weight \(\lambda_{\text{train}}\); inference weight \(\lambda_{\text{infer}}\); diffusion schedule (VP, sub-VP, linear); and advanced distillation variants (Rectified Flow, DMD, Consistency Model). Figure~\ref{fig:ablation_study} presents three key ablation studies on the Darcy dataset. Panel (a) shows that increasing the teacher model's sampling steps $N_s$ consistently improves both generative quality and physical alignment, as reflected by lower MMSE, SMSE, and PDE residuals of the distilled student, highlighting the importance of high-fidelity supervision. Panels (b) and (c) examine the impact of the PDE loss weight during distillation and inference, respectively. We observe that across a wide range of weights, all metrics achieve strong performance. In particular, compared with baselines in Table~\ref{tab:downstream_evaluations}, the PDE error is reduced by an order of magnitude. This improvement is consistent with the design goal of enforcing PDE constraints on final generated samples rather than posterior-mean surrogates.

We also explore whether more sophisticated distillation strategies can improve the quality of the student model. As shown in Table~\ref{tab:ablation_more_distillation}, advanced techniques such as Rectified Flow (RF-1, RF-2), Distribution Matching Distillation (DMD), and Consistency Model yield better MMSE and SMSE than our raw method, while maintaining competitive PDE residuals. This indicates that tighter coupling between noise and data trajectories during distillation facilitates noise-data learning. In addition, we analyze the effect of imposing hard constraints during downstream inference. Following the strategy inspired by ECI-sampling, we directly replace the masked entries in the generated sample with observed values before computing the PDE residual. This ensures that the known information is preserved when evaluating physical consistency. As shown by the $-\text{HC}$ variant in Table~\ref{tab:ablation_more_distillation}, removing this hard-constraint replacement significantly degrades the PDE residual errors across all tasks. We also validate our design using other linear diffusion processes in Table~\ref{tab:ablation_more_distillation} (i.e., $\text{VP}$ and $\text{sub-VP}$).

Appendix~\ref{appendix:additional_experiments} provides additional diagnostics for the main design choices. Table~\ref{tab:appendix_component_attribution} separates one-step distillation from final-sample PDE supervision, showing that distillation mainly gives speed while final-sample supervision drives physical consistency; Table~\ref{tab:appendix_final_sample_supervision} confirms the same pattern across RF, DMD, and consistency-model backbones. Table~\ref{tab:appendix_diversity} further shows that PIDDM preserves sample diversity better than vanilla distillation. We also test robustness to teacher quality and architecture in Tables~\ref{tab:appendix_teacher_quality} and~\ref{tab:appendix_architecture}, and report dynamic weighting and latent-optimization stability in Tables~\ref{tab:appendix_dynamic_weighting}--\ref{tab:appendix_latent_variance}.

\section{Conclusion and Limitation}

We propose \textbf{PIDDM}, a lightweight post-hoc distillation framework for physics-constrained diffusion models. Unlike methods that impose PDE constraints on the posterior mean, leading to a Jensen's Gap and a trade-off between quality and constraint satisfaction, PIDDM enforces constraints directly on the final generated sample. This design improves sample-level physical consistency while empirically preserving distributional fidelity. We provide empirical illustrations of the Jensen's Gap in diffusion training and sampling. Experiments show PIDDM improves physical and distributional fidelity across forward, inverse, and partial reconstruction tasks, with robustness to hyperparameter choice. The distilled student model also enables efficient one-step physics simulation for repeated-query and latency-sensitive scientific workflows such as uncertainty quantification, online data assimilation, virtual sensing, and design optimization.\\
\noindent\textbf{Limitations.} PIDDM requires a well-trained teacher model and a reliable PDE residual operator, which can be challenging to obtain. The one-step student's performance may degrade if the teacher is poorly calibrated or lacks diversity. Although performance is robust across a wide range of PDE loss weights, some hyperparameter tuning is still needed to obtain optimal results. Addressing these limitations remains important future work.

\section*{Acknowledgements}
We sincerely thank the reviewers and the area chair for their careful reading, constructive feedback, and thoughtful suggestions, which helped us improve the clarity and presentation of this work.  D. Zou and Y. Zhang are supported in part by NSFC 62306252, Hong Kong ECS award 27309624, Guangdong NSF
2024A1515012444, and the central fund from HKU. P. Wang is supported in part by the University of Macau SRG2025-00043-FST and UMDF-TISF-I/2026/013/FST, and in part by the Macau Science and Technology Development Fund (FDCT) 0091/2025/ITP2.

\section*{Impact Statement}
This work makes physics-constrained diffusion models practical by enforcing PDE constraints post hoc, sidestepping the posterior-mean mismatch in the student objective and enabling fast single-step generation. By improving physical fidelity and reducing compute and tuning burden, PIDDM can accelerate simulation, inverse design, and data completion workflows in engineering and scientific domains.

\bibliography{main}

@inproceedings{song2020score,
  title={Score-Based Generative Modeling through Stochastic Differential Equations},
  author={Song, Yang and Sohl-Dickstein, Jascha and Kingma, Diederik P and Kumar, Abhishek and Ermon, Stefano and Poole, Ben},
  booktitle={International Conference on Learning Representations},
  year={2020}
}

@article{ho2020ddpm,
  title={Denoising diffusion probabilistic models},
  author={Ho, Jonathan and Jain, Ajay and Abbeel, Pieter},
  journal={Advances in neural information processing systems},
  volume={33},
  pages={6840--6851},
  year={2020}
}

@inproceedings{rombach2022ldm,
  title={High-resolution image synthesis with latent diffusion models},
  author={Rombach, Robin and Blattmann, Andreas and Lorenz, Dominik and Esser, Patrick and Ommer, Bj{\"o}rn},
  booktitle={Proceedings of the IEEE/CVF conference on computer vision and pattern recognition},
  pages={10684--10695},
  year={2022}
}

@inproceedings{guo2024gradient,
  title={Gradient guidance for diffusion models: An optimization perspective},
  author={Guo, Yingqing and Yuan, Hui and Yang, Yukang and Chen, Minshuo and Wang, Mengdi},
  booktitle={Advances in Neural Information Processing Systems (NeurIPS) 37},
  pages={90736--90770},
  year={2024}
}

@article{kerrigan2023functional,
  title={Functional flow matching},
  author={Kerrigan, Gavin and Migliorini, Giosue and Smyth, Padhraic},
  journal={arXiv preprint arXiv:2305.17209},
  year={2023}
}

@inproceedings{
cheng2025eci,
title={Gradient-Free Generation for Hard-Constrained Systems},
author={Chaoran Cheng and Boran Han and Danielle C. Maddix and Abdul Fatir Ansari and Andrew Stuart and Michael W. Mahoney and Bernie Wang},
booktitle={The Thirteenth International Conference on Learning Representations},
year={2025},
url={https://openreview.net/forum?id=teE4pl9ftK}
}

@inproceedings{
huang2024diffusionpde,
title={Diffusion{PDE}: Generative {PDE}-Solving under Partial Observation},
author={Jiahe Huang and Guandao Yang and Zichen Wang and Jeong Joon Park},
booktitle={The Thirty-eighth Annual Conference on Neural Information Processing Systems},
year={2024},
url={https://openreview.net/forum?id=z0I2SbjN0R}
}

@inproceedings{
bastek2025pdim,
title={Physics-Informed Diffusion Models},
author={Jan-Hendrik Bastek and WaiChing Sun and Dennis Kochmann},
booktitle={The Thirteenth International Conference on Learning Representations},
year={2025},
url={https://openreview.net/forum?id=tpYeermigp}
}

@misc{jacobsen2024cocogen,
      title={CoCoGen: Physically-Consistent and Conditioned Score-based Generative Models for Forward and Inverse Problems}, 
      author={Christian Jacobsen and Yilin Zhuang and Karthik Duraisamy},
      year={2024},
      eprint={2312.10527},
      archivePrefix={arXiv},
      primaryClass={cs.LG},
      url={https://arxiv.org/abs/2312.10527}, 
}

@article{shu2023physicscfg,
  title={A physics-informed diffusion model for high-fidelity flow field reconstruction},
  author={Shu, Dule and Li, Zijie and Farimani, Amir Barati},
  journal={Journal of Computational Physics},
  volume={478},
  pages={111972},
  year={2023},
  publisher={Elsevier}
}

@article{vanilladistillation,
  title={Knowledge distillation in iterative generative models for improved sampling speed},
  author={Luhman, Eric and Luhman, Troy},
  journal={arXiv preprint arXiv:2101.02388},
  year={2021}
}

@inproceedings{
liu2023rectifiedflow,
title={Flow Straight and Fast: Learning to Generate and Transfer Data with Rectified Flow},
author={Xingchao Liu and Chengyue Gong and qiang liu},
booktitle={The Eleventh International Conference on Learning Representations },
year={2023},
url={https://openreview.net/forum?id=XVjTT1nw5z}
}

@misc{song2023consistencymodels,
      title={Consistency Models}, 
      author={Yang Song and Prafulla Dhariwal and Mark Chen and Ilya Sutskever},
      year={2023},
      eprint={2303.01469},
      archivePrefix={arXiv},
      primaryClass={cs.LG},
      url={https://arxiv.org/abs/2303.01469}, 
}

@inproceedings{liu2023instaflow,
  title={Instaflow: One step is enough for high-quality diffusion-based text-to-image generation},
  author={Liu, Xingchao and Zhang, Xiwen and Ma, Jianzhu and Peng, Jian and others},
  booktitle={The Twelfth International Conference on Learning Representations},
  year={2023}
}

@inproceedings{yin2024dmd,
  title={One-step diffusion with distribution matching distillation},
  author={Yin, Tianwei and Gharbi, Micha{\"e}l and Zhang, Richard and Shechtman, Eli and Durand, Fredo and Freeman, William T and Park, Taesung},
  booktitle={Proceedings of the IEEE/CVF conference on computer vision and pattern recognition},
  pages={6613--6623},
  year={2024}
}

@inproceedings{zheng2023fastno,
  title={Fast sampling of diffusion models via operator learning},
  author={Zheng, Hongkai and Nie, Weili and Vahdat, Arash and Azizzadenesheli, Kamyar and Anandkumar, Anima},
  booktitle={International conference on machine learning},
  pages={42390--42402},
  year={2023},
  organization={PMLR}
}

@article{berthelot2023tract,
  title={Tract: Denoising diffusion models with transitive closure time-distillation},
  author={Berthelot, David and Autef, Arnaud and Lin, Jierui and Yap, Dian Ang and Zhai, Shuangfei and Hu, Siyuan and Zheng, Daniel and Talbott, Walter and Gu, Eric},
  journal={arXiv preprint arXiv:2303.04248},
  year={2023}
}

@article{tee2024pid,
  title={Physics informed distillation for diffusion models},
  author={Tee, Joshua Tian Jin and Zhang, Kang and Yoon, Hee Suk and Gowda, Dhananjaya Nagaraja and Kim, Chanwoo and Yoo, Chang D},
  journal={arXiv preprint arXiv:2411.08378},
  year={2024}
}

@article{gao2017jensen,
  title={Bounds on the jensen gap, and implications for mean-concentrated distributions},
  author={Gao, Xiang and Sitharam, Meera and Roitberg, Adrian E},
  journal={arXiv preprint arXiv:1712.05267},
  year={2017}
}

@article{karras2022edm,
  title={Elucidating the design space of diffusion-based generative models},
  author={Karras, Tero and Aittala, Miika and Aila, Timo and Laine, Samuli},
  journal={Advances in neural information processing systems},
  volume={35},
  pages={26565--26577},
  year={2022}
}

@inproceedings{esser2024sd3,
  title={Scaling rectified flow transformers for high-resolution image synthesis},
  author={Esser, Patrick and Kulal, Sumith and Blattmann, Andreas and Entezari, Rahim and M{\"u}ller, Jonas and Saini, Harry and Levi, Yam and Lorenz, Dominik and Sauer, Axel and Boesel, Frederic and others},
  booktitle={Forty-first International Conference on Machine Learning},
  year={2024}
}

@inproceedings{
lipman2023flow,
title={Flow Matching for Generative Modeling},
author={Yaron Lipman and Ricky T. Q. Chen and Heli Ben-Hamu and Maximilian Nickel and Matthew Le},
booktitle={The Eleventh International Conference on Learning Representations },
year={2023},
url={https://openreview.net/forum?id=PqvMRDCJT9t}
}

@inproceedings{
liu2023flow,
title={Flow Straight and Fast: Learning to Generate and Transfer Data with Rectified Flow},
author={Xingchao Liu and Chengyue Gong and qiang liu},
booktitle={The Eleventh International Conference on Learning Representations },
year={2023},
url={https://openreview.net/forum?id=XVjTT1nw5z}
}

@inproceedings{nichol2021improved,
  title={Improved denoising diffusion probabilistic models},
  author={Nichol, Alexander Quinn and Dhariwal, Prafulla},
  booktitle={International conference on machine learning},
  pages={8162--8171},
  year={2021},
  organization={PMLR}
}

@article{maoutsa2020interactingode,
  title={Interacting particle solutions of fokker--planck equations through gradient--log--density estimation},
  author={Maoutsa, Dimitra and Reich, Sebastian and Opper, Manfred},
  journal={Entropy},
  volume={22},
  number={8},
  pages={802},
  year={2020},
  publisher={MDPI}
}

@inproceedings{
song2021ddim,
title={Denoising Diffusion Implicit Models},
author={Jiaming Song and Chenlin Meng and Stefano Ermon},
booktitle={International Conference on Learning Representations},
year={2021},
url={https://openreview.net/forum?id=St1giarCHLP}
}

@article{lu2022dpmdeterminsitc,
  title={Dpm-solver: A fast ode solver for diffusion probabilistic model sampling in around 10 steps},
  author={Lu, Cheng and Zhou, Yuhao and Bao, Fan and Chen, Jianfei and Li, Chongxuan and Zhu, Jun},
  journal={Advances in Neural Information Processing Systems},
  volume={35},
  pages={5775--5787},
  year={2022}
}

@article{lu2022dpmplusplusdeterminsitc,
  title={Dpm-solver++: Fast solver for guided sampling of diffusion probabilistic models},
  author={Lu, Cheng and Zhou, Yuhao and Bao, Fan and Chen, Jianfei and Li, Chongxuan and Zhu, Jun},
  journal={arXiv preprint arXiv:2211.01095},
  year={2022}
}

@article{shih2024paralleldeterminsitc,
  title={Parallel sampling of diffusion models},
  author={Shih, Andy and Belkhale, Suneel and Ermon, Stefano and Sadigh, Dorsa and Anari, Nima},
  journal={Advances in Neural Information Processing Systems},
  volume={36},
  year={2024}
}

@inproceedings{zhou2024fastdeterminsitc,
  title={Fast ode-based sampling for diffusion models in around 5 steps},
  author={Zhou, Zhenyu and Chen, Defang and Wang, Can and Chen, Chun},
  booktitle={Proceedings of the IEEE/CVF Conference on Computer Vision and Pattern Recognition},
  pages={7777--7786},
  year={2024}
}

@misc{kingma2013vae,
  title={Auto-encoding variational bayes},
  author={Kingma, Diederik P and Welling, Max and others},
  year={2013},
  publisher={Banff, Canada}
}

@article{li2020fno,
  title={Fourier neural operator for parametric partial differential equations},
  author={Li, Zongyi and Kovachki, Nikola and Azizzadenesheli, Kamyar and Liu, Burigede and Bhattacharya, Kaushik and Stuart, Andrew and Anandkumar, Anima},
  journal={arXiv preprint arXiv:2010.08895},
  year={2020}
}

@article{li2024pino,
  title={Physics-informed neural operator for learning partial differential equations},
  author={Li, Zongyi and Zheng, Hongkai and Kovachki, Nikola and Jin, David and Chen, Haoxuan and Liu, Burigede and Azizzadenesheli, Kamyar and Anandkumar, Anima},
  journal={ACM/JMS Journal of Data Science},
  volume={1},
  number={3},
  pages={1--27},
  year={2024},
  publisher={ACM New York, NY}
}

@article{anderson1982reversesde,
  title={Reverse-time diffusion equation models},
  author={Anderson, Brian DO},
  journal={Stochastic Processes and their Applications},
  volume={12},
  number={3},
  pages={313--326},
  year={1982},
  publisher={Elsevier}
}

@article{efron2011tweedie,
  title={Tweedie’s formula and selection bias},
  author={Efron, Bradley},
  journal={Journal of the American Statistical Association},
  volume={106},
  number={496},
  pages={1602--1614},
  year={2011},
  publisher={Taylor \& Francis}
}

@article{ben2024dflow,
  title={D-flow: Differentiating through flows for controlled generation},
  author={Ben-Hamu, Heli and Puny, Omri and Gat, Itai and Karrer, Brian and Singer, Uriel and Lipman, Yaron},
  journal={arXiv preprint arXiv:2402.14017},
  year={2024}
}

@inproceedings{guo2024initno_goldennoise,
  title={Initno: Boosting text-to-image diffusion models via initial noise optimization},
  author={Guo, Xiefan and Liu, Jinlin and Cui, Miaomiao and Li, Jiankai and Yang, Hongyu and Huang, Di},
  booktitle={Proceedings of the IEEE/CVF Conference on Computer Vision and Pattern Recognition},
  pages={9380--9389},
  year={2024}
}

@article{zhou2024golden,
  title={Golden noise for diffusion models: A learning framework},
  author={Zhou, Zikai and Shao, Shitong and Bai, Lichen and Xu, Zhiqiang and Han, Bo and Xie, Zeke},
  journal={arXiv preprint arXiv:2411.09502},
  year={2024}
}

@article{wang2024silent,
  title={The Silent Prompt: Initial Noise as Implicit Guidance for Goal-Driven Image Generation},
  author={Wang, Ruoyu and Huang, Huayang and Zhu, Ye and Russakovsky, Olga and Wu, Yu},
  journal={arXiv preprint arXiv:2412.05101},
  year={2024}
}

@inproceedings{mao2024lottery,
  title={The lottery ticket hypothesis in denoising: Towards semantic-driven initialization},
  author={Mao, Jiafeng and Wang, Xueting and Aizawa, Kiyoharu},
  booktitle={European Conference on Computer Vision},
  pages={93--109},
  year={2024},
  organization={Springer}
}

@inproceedings{chen2024find,
  title={FIND: Fine-tuning Initial Noise Distribution with Policy Optimization for Diffusion Models},
  author={Chen, Changgu and Yang, Libing and Yang, Xiaoyan and Chen, Lianggangxu and He, Gaoqi and Wang, Changbo and Li, Yang},
  booktitle={Proceedings of the 32nd ACM International Conference on Multimedia},
  pages={6735--6744},
  year={2024}
}

@book{leveque1992numerical,
  title={Numerical methods for conservation laws},
  author={LeVeque, Randall J and Leveque, Randall J},
  volume={132},
  year={1992},
  publisher={Springer}
}

@inproceedings{hansen2023learning,
  title={Learning physical models that can respect conservation laws},
  author={Hansen, Derek and Maddix, Danielle C and Alizadeh, Shima and Gupta, Gaurav and Mahoney, Michael W},
  booktitle={International Conference on Machine Learning},
  pages={12469--12510},
  year={2023},
  organization={PMLR}
}

@article{mouli2024using,
  title={Using uncertainty quantification to characterize and improve out-of-domain learning for PDEs},
  author={Mouli, S Chandra and Maddix, Danielle C and Alizadeh, Shima and Gupta, Gaurav and Stuart, Andrew and Mahoney, Michael W and Wang, Yuyang},
  journal={arXiv preprint arXiv:2403.10642},
  year={2024}
}

@article{saad2022guiding,
  title={Guiding continuous operator learning through physics-based boundary constraints},
  author={Saad, Nadim and Gupta, Gaurav and Alizadeh, Shima and Maddix, Danielle C},
  journal={arXiv preprint arXiv:2212.07477},
  year={2022}
}

@article{vaswani2017attention,
  title={Attention is all you need},
  author={Vaswani, Ashish and Shazeer, Noam and Parmar, Niki and Uszkoreit, Jakob and Jones, Llion and Gomez, Aidan N and Kaiser, {\L}ukasz and Polosukhin, Illia},
  journal={Advances in neural information processing systems},
  volume={30},
  year={2017}
}

@inproceedings{chung2022dps,
  title={Diffusion Posterior Sampling for General Noisy Inverse Problems},
  author={Chung, Hyungjin and Kim, Jeongsol and McCann, Michael Thompson and Klasky, Marc Louis and Ye, Jong Chul},
  booktitle={The Eleventh International Conference on Learning Representations},
  year={2023},
  url={https://openreview.net/forum?id=OnD9zGAGT0k}
}

@misc{
wang2025resample,
title={Noise Re-sampling for High Fidelity Image Generation},
author={Hao Wang and Weihua Chen and Chenming Li and Wenjian Huang and Jingkai Zhou and Fan Wang and Jianguo Zhang},
year={2025},
howpublished={Submitted to ICLR 2025},
url={https://openreview.net/forum?id=GD4Tlqvwrq}
}

@article{raissi2019pinn,
  title={Physics-informed neural networks: A deep learning framework for solving forward and inverse problems involving nonlinear partial differential equations},
  author={Raissi, Maziar and Perdikaris, Paris and Karniadakis, George E},
  journal={Journal of Computational physics},
  volume={378},
  pages={686--707},
  year={2019},
  publisher={Elsevier}
}

@article{paszke2019pytorch,
  title={Pytorch: An imperative style, high-performance deep learning library},
  author={Paszke, A},
  journal={arXiv preprint arXiv:1912.01703},
  year={2019}
}

@inproceedings{abadi2016tensorflow,
  title={$\{$TensorFlow$\}$: a system for $\{$Large-Scale$\}$ machine learning},
  author={Abadi, Mart{\'\i}n and Barham, Paul and Chen, Jianmin and Chen, Zhifeng and Davis, Andy and Dean, Jeffrey and Devin, Matthieu and Ghemawat, Sanjay and Irving, Geoffrey and Isard, Michael and others},
  booktitle={12th USENIX symposium on operating systems design and implementation (OSDI 16)},
  pages={265--283},
  year={2016}
}

@book{smith1985fem,
  title={Numerical solution of partial differential equations: finite difference methods},
  author={Smith, Gordon D},
  year={1985},
  publisher={Oxford university press}
}

@book{leveque2007fem,
  title={Finite difference methods for ordinary and partial differential equations: steady-state and time-dependent problems},
  author={LeVeque, Randall J},
  year={2007},
  publisher={SIAM}
}

@book{hughes2003finite,
  title     = {The Finite Element Method: Linear Static and Dynamic Finite Element Analysis},
  author    = {Hughes, Thomas J. R.},
  year      = {2000},
  publisher = {Dover Publications},
  address   = {Mineola, NY},
  isbn      = {9780486411811}
}

@article{kingma2014adam,
  title={Adam: A method for stochastic optimization},
  author={Kingma, Diederik P and Ba, Jimmy},
  journal={arXiv preprint arXiv:1412.6980},
  year={2014}
}

@book{davidson2015turbulence,
  title     = {Turbulence: An Introduction for Scientists and Engineers},
  author    = {Davidson, Peter A.},
  year      = {2015},
  publisher = {Oxford University Press}
}

@book{incropera2011heat,
  title     = {Fundamentals of Heat and Mass Transfer},
  author    = {Incropera, Frank P. and DeWitt, David P. and Bergman, Theodore L. and Lavine, Adrienne S.},
  edition   = {7},
  year      = {2011},
  publisher = {John Wiley \& Sons},
  address   = {Hoboken, NJ},
  isbn      = {9780470501979}
}

@book{timoshenko1970elasticity,
  title     = {Theory of Elasticity},
  author    = {Timoshenko, Stephen P. and Goodier, James N.},
  edition   = {3},
  year      = {1970},
  publisher = {McGraw-Hill},
  address   = {New York},
  isbn      = {9780070647206}
}

@book{jackson1998classical,
  title     = {Classical Electrodynamics},
  author    = {Jackson, John David},
  edition   = {3},
  year      = {1998},
  publisher = {John Wiley \& Sons}
}

@book{crank1975diffusion,
  title     = {The Mathematics of Diffusion},
  author    = {Crank, John},
  edition   = {2},
  year      = {1975},
  publisher = {Clarendon Press},
  address   = {Oxford},
  isbn      = {9780198534112}
}

@article{lu2019deeponet,
  title={Deeponet: Learning nonlinear operators for identifying differential equations based on the universal approximation theorem of operators},
  author={Lu, Lu and Jin, Pengzhan and Karniadakis, George Em},
  journal={arXiv preprint arXiv:1910.03193},
  year={2019}
}

@misc{ho2022cfg,
      title={Classifier-Free Diffusion Guidance}, 
      author={Jonathan Ho and Tim Salimans},
      year={2022},
      eprint={2207.12598},
      archivePrefix={arXiv},
      primaryClass={cs.LG},
      url={https://arxiv.org/abs/2207.12598}, 
}

@misc{benchmark,
      title={InverseBench: Benchmarking Plug-and-Play Diffusion Priors for Inverse Problems in Physical Sciences}, 
      author={Hongkai Zheng and Wenda Chu and Bingliang Zhang and Zihui Wu and Austin Wang and Berthy T. Feng and Caifeng Zou and Yu Sun and Nikola Kovachki and Zachary E. Ross and Katherine L. Bouman and Yisong Yue},
      year={2025},
      eprint={2503.11043},
      archivePrefix={arXiv},
      primaryClass={cs.LG},
      url={https://arxiv.org/abs/2503.11043}, 
}

@misc{ddrm,
      title={Denoising Diffusion Restoration Models}, 
      author={Bahjat Kawar and Michael Elad and Stefano Ermon and Jiaming Song},
      year={2022},
      eprint={2201.11793},
      archivePrefix={arXiv},
      primaryClass={eess.IV},
      url={https://arxiv.org/abs/2201.11793}, 
}

@misc{ddnm,
      title={Zero-Shot Image Restoration Using Denoising Diffusion Null-Space Model}, 
      author={Yinhuai Wang and Jiwen Yu and Jian Zhang},
      year={2022},
      eprint={2212.00490},
      archivePrefix={arXiv},
      primaryClass={cs.CV},
      url={https://arxiv.org/abs/2212.00490}, 
}

@misc{diffpir,
      title={Denoising Diffusion Models for Plug-and-Play Image Restoration}, 
      author={Yuanzhi Zhu and Kai Zhang and Jingyun Liang and Jiezhang Cao and Bihan Wen and Radu Timofte and Luc Van Gool},
      year={2023},
      eprint={2305.08995},
      archivePrefix={arXiv},
      primaryClass={cs.CV},
      url={https://arxiv.org/abs/2305.08995}, 
}

@inproceedings{daps,
   title={Improving Diffusion Inverse Problem Solving with Decoupled Noise Annealing},
   url={http://dx.doi.org/10.1109/CVPR52734.2025.01946},
   DOI={10.1109/cvpr52734.2025.01946},
   booktitle={2025 IEEE/CVF Conference on Computer Vision and Pattern Recognition (CVPR)},
   publisher={IEEE},
   author={Zhang, Bingliang and Chu, Wenda and Berner, Julius and Meng, Chenlin and Anandkumar, Anima and Song, Yang},
   year={2025},
   month=jun, pages={20895–20905} }

@InProceedings{LGD,
  title = 	 {Loss-Guided Diffusion Models for Plug-and-Play Controllable Generation},
  author =       {Song, Jiaming and Zhang, Qinsheng and Yin, Hongxu and Mardani, Morteza and Liu, Ming-Yu and Kautz, Jan and Chen, Yongxin and Vahdat, Arash},
  booktitle = 	 {Proceedings of the 40th International Conference on Machine Learning},
  pages = 	 {32483--32498},
  year = 	 {2023},
  editor = 	 {Krause, Andreas and Brunskill, Emma and Cho, Kyunghyun and Engelhardt, Barbara and Sabato, Sivan and Scarlett, Jonathan},
  volume = 	 {202},
  series = 	 {Proceedings of Machine Learning Research},
  month = 	 {23--29 Jul},
  publisher =    {PMLR},
  url = 	 {https://proceedings.mlr.press/v202/song23k.html}
}

@misc{DPG,
      title={Solving General Noisy Inverse Problem via Posterior Sampling: A Policy Gradient Viewpoint}, 
      author={Haoyue Tang and Tian Xie and Aosong Feng and Hanyu Wang and Chenyang Zhang and Yang Bai},
      year={2024},
      eprint={2403.10585},
      archivePrefix={arXiv},
      primaryClass={eess.IV},
      url={https://arxiv.org/abs/2403.10585}, 
}

@inproceedings{
fps,
title={Diffusion Posterior Sampling for Linear Inverse Problem Solving: A Filtering Perspective},
author={Zehao Dou and Yang Song},
booktitle={The Twelfth International Conference on Learning Representations},
year={2024},
url={https://openreview.net/forum?id=tplXNcHZs1}
}

@inproceedings{
mcgdiff,
title={Monte Carlo Guided Denoising Diffusion Models for Bayesian Linear Inverse Problems},
author={Gabriel Cardoso and Yazid Janati {El Idrissi} and Sylvain {Le Corff} and Eric Moulines},
booktitle={The Twelfth International Conference on Learning Representations},
year={2024},
url={https://openreview.net/forum?id=nHESwXvxWK}
}

@misc{scg,
      title={Symbolic Music Generation with Non-Differentiable Rule Guided Diffusion}, 
      author={Yujia Huang and Adishree Ghatare and Yuanzhe Liu and Ziniu Hu and Qinsheng Zhang and Chandramouli S Sastry and Siddharth Gururani and Sageev Oore and Yisong Yue},
      year={2024},
      eprint={2402.14285},
      archivePrefix={arXiv},
      primaryClass={cs.SD},
      url={https://arxiv.org/abs/2402.14285}, 
}

@misc{pnp-dm,
      title={Principled Probabilistic Imaging using Diffusion Models as Plug-and-Play Priors}, 
      author={Zihui Wu and Yu Sun and Yifan Chen and Bingliang Zhang and Yisong Yue and Katherine L. Bouman},
      year={2024},
      eprint={2405.18782},
      archivePrefix={arXiv},
      primaryClass={eess.IV},
      url={https://arxiv.org/abs/2405.18782}, 
}

@misc{
dcdpm,
title={{DC}-{DPM}: A Divide-and-Conquer Approach for Diffusion Reverse Process},
author={Yue-Jiang Dong and Hubery Yin and Fangyikang Wang and Yizhe Zhao and Chao Zhang and Chen Li and Song-Hai Zhang},
year={2024},
howpublished={ICLR 2025 Conference Withdrawn Submission},
url={https://openreview.net/forum?id=VbAxCwV2e3}
}

@misc{midpoint,
      title={Variational Diffusion Posterior Sampling with Midpoint Guidance}, 
      author={Badr Moufad and Yazid Janati and Lisa Bedin and Alain Durmus and Randal Douc and Eric Moulines and Jimmy Olsson},
      year={2024},
      eprint={2410.09945},
      archivePrefix={arXiv},
      primaryClass={stat.ML},
      url={https://arxiv.org/abs/2410.09945}, 
}
\bibliographystyle{plainnat}
\newpage
\appendix

\onecolumn
\section{Related Work}
\subsection{Diffusion Models}
Diffusion models \cite{song2020score, ho2020ddpm, karras2022edm} learn a score function, $\nabla \log p(\vx_t)$, to reverse a predefined diffusion process, typically of the form $\vx_t = \vx_0 + \sigma_t \boldsymbol{\varepsilon}$. A key characteristic of diffusion models is that sampling requires iteratively reversing this process over a sequence of timesteps. This iterative nature presents a challenge for controlled generation: to guide the sampling trajectory effectively, we often need to first estimate the current denoised target $x_0$ in order to determine the correct guidance direction. In other words, \textit{to decide how to get there, we must first understand where we are}. However, obtaining this information through full iterative sampling is computationally expensive and often impractical in an optimization regime.

A practical workaround is to leverage an implicit one-step data estimate provided by diffusion models via Tweedie's formula \cite{efron2011tweedie}, which requires only a single network forward pass:
\begin{equation*}
\hat{\vx}_0\approx \mathbb{E}[\vx_0|\vx_t]= \vx_t+\sigma_t^2\nabla \log p(\vx_t),
\end{equation*}
where $\hat{\vx}_0$ denotes a one-step denoised estimate of the clean sample from $\vx_t$. This approximation improves as $t \to0$.
Although this posterior mean $\mathbb{E}[\vx_0 \mid \vx_t]$ is not theoretically equivalent to the final sample obtained after full denoising, in practice, this estimate serves as a useful proxy for the underlying data and enables approximate guidance for controlled generation, without the need to complete the entire sampling trajectory.

\subsection{Constrained Generation for PDE Systems} \label{Appendix:related-work:constrained-generation}

Diffusion models have demonstrated strong potential for physical-constraint applications due to their generative nature. This generative capability naturally supports the basic task of simulating physical data and also extends to downstream applications such as reconstruction from partial observations and solving both forward and inverse problems. However, many scientific tasks require strict adherence to physical laws, often expressed as PDE constraints on the data. These constraints, applied at the sample level \(\vx\), are not easily enforced within diffusion models, which are trained to model the data distribution \(p(\vx)\). To address this, prior works have proposed three main strategies for incorporating physical constraints into diffusion models.

\textbf{Training-time Loss Injection.} PG-Diffusion~\cite{shu2023physicscfg} employs Classifier-Free Guidance (CFG), where a conditional diffusion model is trained using the PDE residual error as a conditioning input. However, CFG is well known to suffer from theoretical inconsistencies---specifically, the interpolated conditional score function does not match the true conditional score---which limits its suitability for enforcing precise physical constraints. To avoid this issue, PIDM~\cite{bastek2025pdim} introduces a loss term based on the residual evaluated at the posterior mean, \(\mathbb{E}[\vx_0 \mid \vx_t]\). While this approach avoids the theoretical pitfalls of CFG, the constraint is still not imposed on the actual sample \(\vx_0\), inheriting the Jensen-gap issue originally discussed in diffusion posterior sampling for inverse problems~\cite{chung2022dps}.

\textbf{Sampling-time Guidance.} Diffusion Posterior Sampling (DPS), used in DiffusionPDE~\cite{huang2024diffusionpde} and CoCoGen~\cite{jacobsen2024cocogen}, applies guidance during each sampling step by using the gradient of the PDE residual evaluated on the posterior mean \(\mathbb{E}[\bm x_0 \mid \bm x_t]\). Therefore, they inherit the Jensen’s Gap issue, as the guidance operates on an estimate of the final sample rather than the sample itself. Moreover, DPS assumes that the residual error follows a Gaussian distribution—a condition that may not hold in real-world PDE systems. Meanwhile, to support hard constraints, ECI-sampling~\cite{cheng2025eci} directly modifies the posterior mean \(\mathbb{E}[\vx_0 \mid \vx_t]\) using known boundary conditions.

\textbf{Noise Prompting.} Another stream of research—often called \emph{noise prompting} or \emph{golden-noise optimization}—directly tunes the \emph{initial} noise so that the resulting sample satisfies a target constraint~\cite{ben2024dflow,guo2024initno_goldennoise,zhou2024golden,wang2024silent,mao2024lottery,chen2024find}.
In the physics domain, this idea is used to minimize the true PDE residual \(R(\vx)\) evaluated on the \emph{final} sample, rather than the surrogate residual \(R(\mathbb{E}[\vx_0 \mid \vx_t])\). Because the constraint is imposed on the actual output, noise prompting avoids posterior-mean surrogate constraints and therefore serves as a strong baseline in ECI-sampling~\cite{cheng2025eci} and PIDM~\cite{bastek2025pdim}. The main drawback is efficiency: optimizing the noise requires backpropagating through the entire sampling trajectory, which is computationally expensive and prone to gradient instability.

Recently, diffusion-based techniques for solving image inverse problems have demonstrated competitive performance \cite{benchmark}. However, a common limitation is that they rely on the posterior mean, i.e., $\mathbb{E}[\vx_0 \mid \vx_t]$, as a surrogate for the true posterior. For example, DDRM \cite{ddrm} and DDNM \cite{ddnm} exploit singular value decomposition (SVD) and pseudo-inverse operations to fill in the missing components of $\mathbb{E}[\vx_0 \mid \vx_t]$ during sampling, which is conceptually similar to ECI-sampling. Likewise, methods such as DiffPIR \cite{diffpir} and DAPS \cite{daps} optimize or run Langevin MCMC updates on the posterior mean in order to enforce observation consistency. Another line of work approximates the likelihood $p(\vy \mid \vx_t)$. For instance, DPS \cite{chung2022dps} treats $p(\vx_0 \mid \vx_t)$ as a point mass centered at $\mathbb{E}[\vx_0 \mid \vx_t]$, while LGD \cite{LGD} and DPG \cite{DPG} use a Gaussian distribution with mean $\mathbb{E}[\vx_0 \mid \vx_t]$ for approximation. To reduce this approximation error, some methods trade off computational cost: Monte Carlo–based approaches \cite{fps, mcgdiff, scg} and variational inference–based approaches \cite{pnp-dm} avoid the direct mean approximation, but they either require simulating a large number of samples or solving intermediate optimization problems during sampling, both of which are computationally expensive. Finally, there are methods that explicitly aim to reduce the Jensen’s Gap by modifying the sampling dynamics. Examples include mid-point schemes \cite{midpoint} and user-defined intermediate potentials \cite{dcdpm}. While these approaches can shorten the gap, they still rely on $\mathbb{E}[\vx_0 \mid \vx_t]$ for posterior estimation, and moreover, they often incur high computational cost due to additional variational inference or Langevin MCMC steps.

\subsection{Distillation of Diffusion Models}

Sampling in diffusion models involves integrating through a reverse diffusion process, which is computationally expensive. Even with the aid of high-order ODE solvers \cite{maoutsa2020interactingode, song2021ddim, lu2022dpmdeterminsitc, lu2022dpmplusplusdeterminsitc, zhou2024fastdeterminsitc}, parallel sampling \cite{shih2024paralleldeterminsitc}, and better training schedules~\cite{karras2022edm, nichol2021improved, liu2023flow, liu2023rectifiedflow, liu2023instaflow}, the process remains iterative and typically requires hundreds of network forward passes. To alleviate this inefficiency, distillation-based methods have been developed to enable one-step generation by leveraging the deterministic nature of samplers (e.g., DDIM), where the noise–data pairs become fixed. The most basic formulation, Knowledge Distillation~\cite{vanilladistillation}, trains a student model to replicate the teacher's deterministic noise-to-data mapping. However, subsequent studies have shown that directly learning this raw mapping is challenging for neural networks, as the high curvature of sampling trajectories often yields noise–data pairs that are distant in Euclidean space, making the regression task ill-conditioned and hard to generalize.

To address this, recent research has proposed three complementary strategies. (1) Noise–data coupling refinement: Rectified Flow \cite{liu2023rectifiedflow} distills the sampling process into a structure approximating optimal transport, where the learned mapping corresponds to minimal-cost trajectories between noise and data. InstaFlow \cite{liu2023instaflow} further demonstrates that such near-optimal-transport couplings significantly ease the learning process for student models. (2) Distribution-level distillation: Rather than matching individual noise–data pairs, DMD \cite{yin2024dmd} trains the student via score-matching losses that align the overall data distributions, thereby bypassing the need to regress complex mappings directly. (3) Trajectory distillation: Instead of only supervising on initial ($\vx_T$) and final ($\vx_0$) states, this approach provides supervision at intermediate states $x_t$ along the ODE trajectory \cite{berthelot2023tract, zheng2023fastno, song2023consistencymodels, tee2024pid}. This decomposition allows the student model to learn the generative process in a piecewise manner, which improves stability and sample fidelity. We note that among existing approaches, Physics-Informed Distillation (PID)~\cite{tee2024pid} has a name similar to our method's but differs fundamentally in both objective and methodology. Specifically, PID distills ODE trajectories from teacher models using a PINN-like strategy, whereas our method distills diffusion models for PDE-constrained generation by applying physical supervision directly to the final samples.

\section{Mixture-of-Gaussians (MoG) Dataset}

To study the sampling-time behavior of constrained diffusion models, we design a synthetic 2D Mixture-of-Gaussians (MoG) dataset with analytical score functions. Each sample $x = (x_1, x_2) \in \mathbb{R}^2$ consists of a data dimension $x_1$ and a fixed latent code $x_2$ that serves as a hard constraint.

Specifically, we define a mixture model where $x_1$ is sampled from a Gaussian mixture conditioned on the latent code $z \in \{-1, +1\}$, and $x_2$ is deterministically set to $z$. The full distribution is:
\begin{equation}
x_2 = z \in \{-1, +1\}, \quad x_1 \sim \mathcal{N}(\mu_z, \sigma^2),
\end{equation}
with $\mu_{-1} = -1$, $\mu_{+1} = +1$, and fixed standard deviation $\sigma = 0.2$. The full 2D data point is thus given by:
\begin{equation}
x = \begin{bmatrix} x_1 \\ x_2 \end{bmatrix}, \quad \text{with } x_1 \sim \mathcal{N}(\mu_{x_2}, \sigma^2), \quad x_2 \in \{-1, +1\}.
\end{equation}

The resulting joint density $p(x)$ is a mixture of two Gaussians supported on parallel horizontal lines:
\begin{equation}
p(x) = \frac{1}{2} \, \mathcal{N}(x_1; -1, \sigma^2) \cdot \delta(x_2 + 1) + \frac{1}{2} \, \mathcal{N}(x_1; +1, \sigma^2) \cdot \delta(x_2 - 1),
\end{equation}
where $\delta(\cdot)$ denotes the Dirac delta function. In our experiment comparing DPS in Sec.~\ref{sec:jensengap-demo}, we tune the weight of DPS guidance to be 0.035, as it gives stable performance.

\subsection{Derivation of Score Function of the MoG Dataset}\label{appendix:mog-score-calculate}

MoG distributions admit analytical diffusion objectives. Specifically, consider a MoG with the form:
\begin{equation*}
\bm{x}_{0} \sim \frac{1}{K} \sum_{k=1}^K  \mathcal{N}(\vmu_k, \sigma_k^2\cdot \mI),
\end{equation*}
where $K$ is the number of Gaussian components, $\vmu_k$ and $\sigma_k^2$ are the means and variances of the Gaussian components, respectively.
Suppose the solution of the diffusion process follows:
\begin{equation*}
\vx_t = \alpha_t\vx_0+ \sigma_t\cdot \xi \quad \mathrm{where} \quad \xi \sim \mathcal{N}(0, \bm{I}).
\end{equation*}
Since $\vx_0$ and $\xi$ are both sampled from Gaussian distributions, their linear combination $\vx_t$ also forms a Gaussian distribution, i.e.,
\begin{equation*}
    \bm{x}_t \sim \frac{1}{K}\sum_{k=1}^K  \mathcal{N}(\alpha_t\vmu_k , (\sigma_k^2\alpha_t^2+ \sigma_t^2)\cdot \bm{I}).
\end{equation*}

Let \(p_i(\vx_t)=\mathcal{N}(\vx_t;\alpha_t\vmu_i,(\sigma_i^2\alpha_t^2+\sigma_t^2)\bm I)\). Then,
\begin{equation*}
\begin{aligned}
\nabla p_t(\vx_t)
&= \frac{1}{K}\sum_{i=1}^{K} \nabla_{\vx_t} p_i(\vx_t)\\
&= \frac{1}{K}\sum_{i=1}^{K} p_i(\vx_t)\cdot
\frac{-(\vx_t-\alpha_t\vmu_i)}{\sigma_i^2\alpha_t^2+\sigma_t^2}.
\end{aligned}
\end{equation*}

We can also calculate the score of $\vx_t$, i.e.,
\begin{equation*}
\begin{aligned}
\nabla \log p_t(\vx_t)
=\frac{\nabla p_t(\vx_t)}{p_t(\vx_t)}
= \frac{\sum_{i=1}^{K} p_i(\vx_t)\cdot
\left(\frac{-(\vx_t-\alpha_t\vmu_i)}{\sigma_i^2\alpha_t^2+\sigma_t^2}\right)}
{\sum_{i=1}^{K} p_i(\vx_t)}.
\end{aligned}
\end{equation*}
\subsection{Deviation of Velocity Field of Reverse ODE and DPS}\label{appendix:devi-velocity}

Diffusion models define a forward diffusion process to perturb the data distribution $p_{\rm data}$ to a Gaussian distribution. Formally, the diffusion process is an Itô SDE $\mathrm{d}\vx_t=f(\vx_t)+g(t)\mathrm{d}\mathbf{w}$, where $\mathrm{d}\mathbf{w}$ is the Brownian motion and $t$ flows forward from $0$ to $T$. The solution of this diffusion process gives a transition distribution $p_t(\vx_t|\vx_0)=\cN(\vx_t|\alpha_t\vx_0,\sigma^2_t\bm{I})$, where $\alpha_t=\exp\left({\int_0^t f(s)ds}\right)$ and $\sigma_t^2=1-\exp\left({-\int_0^t g(s)^2ds}\right)$. For the linear interpolation used in our experiments, $\alpha_t=1-t$ and $\sigma_t=t$. To sample from the diffusion model, a typical approach is to apply a reverse-time SDE that reverses the diffusion process \cite{anderson1982reversesde}:
\begin{equation*}
    \mathrm{d}\vx_t=\left(f(\vx_t)-g(t)^2\nabla_{\vx_t} \log p_t(\vx_t)\right)\mathrm{d}t+g(t)\mathrm{d}\bar{\mathbf{w}},
\end{equation*}
where $\mathrm{d}\bar{\mathbf{w}}$ is the Brownian motion and $t$ flows backward from $T$ to $0$. For all reverse-time SDEs, there exist corresponding deterministic processes that share the same density evolution, i.e., $\{p_t(\bm x_t)\}_{t=0}^T$ \cite{song2020score}. Specifically, this deterministic process follows an ODE:
\begin{equation*}
    \mathrm{d}\vx_t=\left(f(\vx_t)-\frac{1}{2}g(t)^2\nabla_{\vx_t} \log p_t(\vx_t)\right)\mathrm{d}t,
\end{equation*}
 where $t$ flows backward from $T$ to $0$. The deterministic process defines a velocity field,
\begin{equation*}
 v_{\text{GT}}(\vx,t)=[f(\vx_t)-\frac{1}{2}g(t)^2\nabla_{\vx_t} \log p_t(\vx_t)].
\end{equation*}
Here, we also define the velocity field by $v(\vx_t,t)=f(\vx_t)-\frac{1}{2}g(t)^2\nabla_{\vx_t} \log p_t(\vx_t)$.

The posterior mean can be estimated from the score by:

\begin{equation*}
 \mathbb{E}[\vx_0|\vx_t]=\frac{\vx_t + \sigma_t^2 \nabla \log p_t(\vx_t)}{\alpha_t}.
\end{equation*}

The posterior mean can also be estimated from the velocity field by:
\begin{equation*}
 \mathbb{E}[\vx_0|\vx_t]=\frac{\dot{\sigma}_t \vx_t - \sigma_t v(\vx_t,t)}{\alpha_t\dot{\sigma}_t-\sigma_t\dot{\alpha}_t}.
\end{equation*}



\subsection{Theoretical Explanation of the Effectiveness of Distillation}\label{app:pf-sampling}

In this subsection, we provide a theoretical explanation for why the post-hoc distillation stage improves sampling efficiency compared to both standard diffusion models and models with low-rank guidance. Here, we consider a target distribution $\bm x_0 \sim \mathcal{N}(\bm 0, \bm \Lambda)$, where $\bm \Lambda = \mathrm{diag}(\lambda_1,\dots,\lambda_{d-1},0)$ is diagonal without loss of generality and $1 = \lambda_1 \ge \cdots \ge \lambda_{d-1} > 0$. The zero eigenvalue $\lambda_d=0$ is introduced to simulate the hard equality constraint on the data, and the corresponding eigendirection $\bm e_d$ is referred to as the constraint direction.

We consider the variance-preserving SDE with the following forward process:
\begin{equation*}
    d\bm x_t = -\frac{1}{2}\beta \bm x_t dt + \sqrt{\beta}d\bm w_t,\quad  0 \le t \le T,
\end{equation*}
where $\beta > 0$ is a noise schedule and $\bm w_t$ is a standard Wiener process \cite{song2020score}. This implies
\begin{align*}
    \bm x_t = \alpha_t \bm x_0 + \sigma_t\bm \epsilon,\ \text{where}\ \alpha_t = \exp\left( -\frac{1}{2}\int_0^t \beta dy \right),\ \sigma_t = \sqrt{1-\alpha_t^2},\ \text{and}\ \bm \epsilon \sim \mathcal{N}(\bm 0,\bm{I}).
\end{align*}
One can easily verify that the score function is
\begin{align*}
    \nabla \log p_t(\bm x) = -\left(\alpha_t^2\bm \Lambda + \sigma_t^2\bm{I}\right)^{-1}\bm x.
\end{align*}
For ease of exposition, let $\bm \theta^* := -\mathrm{diag}\left(\alpha_t^2\bm \Lambda + \sigma_t^2\bm{I}\right)^{-1}$ denote the diagonal entries and $ \nabla \log p_t(\bm x)=\mathrm{diag}(\bm \theta^* )\bm x$.  In particular, along the constrained direction $\bm e_d$, we have $\theta_d^*(t)=-\sigma_t^{-2}$, whose magnitude diverges as $\sigma_t^2\rightarrow 0$.

\paragraph{Score network.} To learn the score function at each time $t$, we parameterize the score network as $s_{\bm \theta}(\bm x, t) = \mathrm{diag}(\bm \theta)\bm x$ and consider the denoising score matching loss:
\begin{align*}
    \E_{\bm x_0,\bm \epsilon \sim \mathcal{N}(\bm 0,\bm{I})}\left[ \left\|   s_{\bm \theta}(\bm x_t,t) - \nabla \log p_t(\bm x_t) \right\|^2\right].
\end{align*}
Using the above loss, we learn the score function $\hat{s}_{\bm \theta}$ from training samples $\{\bm x^{(i)}\}_{i=1}^M \sim \mathcal{N}(\bm 0, \bm \Lambda)$ by solving the following problem:
\begin{align}\label{eq:Lt}
        \mathcal{L}_t(\bm \theta) & := \frac{1}{M}\sum_{i=1}^M \E_{\bm \epsilon \sim \mathcal{N}(\bm 0, \bm I)}\left[ \left\| \bm s_{\bm \theta}(\alpha_t \bm x_0^{(i)} + \sigma_t \bm \epsilon,t) - \nabla \log p(\bm x_t \mid \bm x_0^{(i)}) \right\|^2\right] \notag\\
    & = \frac{1}{M}\sum_{i=1}^M  \E_{\bm \epsilon \sim \mathcal{N}(\bm 0, \bm I)}\left[ \left\|\mathrm{diag}(\bm \theta)(\alpha_t \bm x_0^{(i)} + \sigma_t \bm \epsilon) + \frac{1}{\sigma_t}\bm \epsilon\right\|^2  \right] \notag\\
        & = \frac{1}{M}\sum_{i=1}^M  \E_{\bm \epsilon \sim \mathcal{N}(\bm 0, \bm I)}\left[ \left\|\alpha_t\mathrm{diag}(\bm \theta)\bm x_0^{(i)} + \left(\sigma_t\mathrm{diag}(\bm \theta) + \frac{1}{\sigma_t}\bm I \right)\bm \epsilon\right\|^2 \right] \notag\\
    & = \frac{\alpha_t^2}{M}\sum_{i=1}^M\left\|\mathrm{diag}(\bm \theta)\bm x_0^{(i)}\right\|^2 + \left\|\sigma_t\mathrm{diag}(\bm \theta) + \frac{1}{\sigma_t}\bm I \right\|_F^2.
\end{align}
Moreover, the effective score function of {\em diffusion models with guidance} \cite{guo2024gradient} is
    \begin{align*}
        s^{\rm g}_{\bm \theta} = s_{\bm \theta}(\bm x_t, t) - \gamma \left\langle \frac{\bm x_t - \sigma_t s_{\bm \theta}(\bm x_t, t)}{\alpha_t}, \bm e_d\right\rangle  \bm e_d,
    \end{align*}
    where the guidance is designed by enforcing the predicted clean data $\hat {\bm x}_{0|t} =\frac{\bm x_t - \sigma_t s_{\bm \theta}(\bm x_t, t)}{\alpha_t} $  to have a small constraint-satisfaction loss $(\hat {\bm x}_{0|t}^\top\mathbf e_d)^2$, and $\gamma$ is the guidance strength. Based on $s_{\bm \theta}(\bm x, t) = \mathrm{diag}(\bm \theta)\bm x$, one can express the guided score as $s^{\rm g}_{\bm \theta} = \mathrm{diag}(\bm \theta^{g})\bm x$.

Then, let $\hat{\bm x}_0(\boldsymbol{\epsilon})$ be the data generated from the input random variable $\boldsymbol{\epsilon}\sim \mathcal{N}(0,\bm I)$ through the ODE solver using the trained score function $s_{\bm \theta}$. The student network $\bm v_{\bm w}(\boldsymbol{\epsilon})=\mathrm{diag}(\bm w)\boldsymbol{\epsilon}$ is trained according to \eqref{eq:distillation_loss}, i.e.,
\begin{align*}
 \mathcal{L}_{\mathrm{distill}}(\bm w)= \mathbb E_{\boldsymbol{\epsilon}}[\|\bm v_{\bm w}(\boldsymbol{\epsilon})-\hat{\bm x}_0(\boldsymbol{\epsilon})\|_2^2] + \lambda (\bm v_{\bm w}(\boldsymbol{\epsilon})^\top \bm e_d)^2.
\end{align*}

\paragraph{Constrained direction may be imperfectly learned via vanilla diffusion models and diffusion models with guidance.} Based on the above setup, we can analyze the learning ability of these methods in terms of the hard constraint on the final coordinate. In particular, we will show that even in the simple linearized setting, learning the constrained coordinate $\theta_d^*(t)$ will be slow at small noise levels, i.e., $t\ll 1$.

\begin{theorem}\label{thm:sampling}
    Suppose that the number of training samples $M$ is sufficiently large and we apply gradient descent to solve the empirical loss in Problem \eqref{eq:Lt}, starting from $\bm \theta^0 = \bm{1}$ and using step size $\eta\sim \Theta(1)$ for $\ell$ iterations. Then, with high probability, for the constrained coordinate \(d\), the iterates satisfy
\[
\theta_d^{\ell} - \theta_d^*(t) = \bigl(1-2\eta\sigma_t^2\bigr)^{\ell}\bigl(\theta_d^{0}-\theta_d^*(t)\bigr),
\qquad \ell=0,1,2,\dots,
\]
and consequently, for any regime with \(\ell\sigma_t^2 \ll 1\),
\[
\frac{\lvert \theta_d^{\ell} - \theta_d^*(t)\rvert}{\lvert \theta_d^*(t)\rvert}
= 1 - \mathcal{O}(\ell\sigma_t^2),
\]
i.e., the learned score coefficient in the constrained direction remains a \emph{constant-factor} away from the optimum unless \(\ell=\Omega(\sigma_t^{-2})\).
For diffusion models with guidance, we have
\[
\theta_d^{\ell,\mathrm{g}}
=
\left(1+\frac{\gamma\sigma_t}{\alpha_t}\right)\theta_d^\ell
-
\frac{\gamma}{\alpha_t}.
\]
\end{theorem}

\begin{proof}
    At time $t$, the considered denoising score matching loss \eqref{eq:Lt} over the training samples $\{\bm x^{(i)}\}_{i=1}^M \sim \mathcal{N}(\bm 0, \bm \Sigma)$ is as follows:
\begin{align*}
    \mathcal{L}_t(\bm \theta)
 = \frac{\alpha_t^2}{M}\sum_{i=1}^M\left\|\mathrm{diag}(\bm \theta)\bm x^{(i)}\right\|^2 + \left\|\sigma_t\mathrm{diag}(\bm \theta) + \frac{1}{\sigma_t}\bm{I} \right\|_F^2.
\end{align*}
For ease of exposition, let
\begin{align}
    \hat{\lambda}_j := \frac{1}{M}\sum_{i=1}^M \left( x_{j}^{(i)}\right)^2.
\end{align}
Note that since $\bm \Sigma = \mathrm{diag}(\lambda_1,\dots,\lambda_{d-1},0)$ with $1 = \lambda_1 \ge \dots \ge \lambda_{d-1} > 0$, it holds with high probability that $\hat{\lambda}_j = \left(1 - o(1)\right)\lambda_j$ for all $j=1,\dots,d-1$ and $\hat{\lambda}_d = 0$ almost surely.
Then, we have
\begin{align*}
      \mathcal{L}_t(\bm \theta) = \sum_{j=1}^d \left( \alpha_t^2 \hat{\lambda}_j \theta_j^2 + \left( \sigma_t  \theta_j + \frac{1}{\sigma_t} \right)^2 \right).
\end{align*}
With initialization $\bm \theta^0 = \bm 1_d$, the gradient descent update is
\begin{align*}
\bm \theta^{\ell+1} = \bm \theta^\ell - \eta \nabla \mathcal{L}_t(\bm \theta^\ell),\quad \ell = 0,1,2,\dots.
\end{align*}
This is equivalent to
\begin{align*}
 \theta_j^{\ell+1} & = \theta_j^{\ell} - 2\eta\left( \left(\alpha_t^2\hat{\lambda}_j + \sigma_t^2\right)\theta_j^\ell + 1 \right)\\
 & = \left( 1 - 2\eta\left( \alpha_t^2 \hat{\lambda}_j + \sigma_t^2 \right) \right) \theta_j^\ell - 2\eta,\ \forall j = 1,\dots,d.
\end{align*}
Obviously, the optimal solution is
\begin{align*}
     \theta_j^* = - \frac{1}{\alpha_t^2\hat{\lambda}_j  + \sigma_t^2},\ \forall j = 1,\dots,d.
\end{align*}
Then, we have
\begin{align*}
    \theta_j^{\ell+1} - \theta_j^* = \left( 1 - 2\eta\left( \alpha_t^2 \hat{\lambda}_j + \sigma_t^2 \right) \right)\left( \theta_j^{\ell} - \theta_j^* \right).
\end{align*}
To guarantee monotonically decreasing for $j=1$, we have
\begin{align*}
 \eta < \frac{1}{2\left(\alpha_t^2\hat{\lambda}_{1} + \sigma_t^2\right)}  \approx \frac{1}{2},
\end{align*}
where the last inequality follows from $\hat{\lambda}_1 \approx 1$. Then, we set $\eta = 1/4$.  Next, we have
\begin{align*}
     \theta_d^{\ell+1} - \theta_d^* & = \left( 1 - 2\eta \sigma_t^2 \right)\left( \theta_d^{\ell} - \theta_d^* \right)
      = \left( 1 - \frac{\sigma_t^2 }{2} \right)\left( \theta_d^{\ell} - \theta_d^* \right).
\end{align*}
This, together with $\theta_d^0 = 1$, implies for all $\ell=1,2,\dots$,
\begin{align*}
     \theta_d^{\ell} - \theta_d^*  = \left( 1 - \frac{\sigma_t^2 }{2}\right)^{\ell} \left(1 - \theta_d^*\right).
\end{align*}
Then by the approximation \((1-2\eta\sigma_t^2)^\ell = 1 - \mathcal{O}(\ell\sigma_t^2)\) when \(\ell\sigma_t^2\ll 1\), we can obtain
\[
\frac{\lvert \theta_d^{\ell} - \theta_d^*(t)\rvert}{\lvert \theta_d^*(t)\rvert}
= 1 - \mathcal{O}(\ell\sigma_t^2).
\]

Thus, the learned score function for standard diffusion models, $\hat{s}_{\bm \theta}(\bm x,t) =  \mathrm{diag}(\bm \theta^\ell)\bm x$, can remain inaccurate in the constrained coordinate.

Next, we consider the effective score function of diffusion models with guidance:
\begin{align*}
     \hat{s}^{\rm g}_{\bm \theta}(\bm x,t) & = \hat{s}_{\bm \theta}(\bm x, t) - \gamma \left\langle \frac{\bm x - \sigma_t \hat{s}_{\bm \theta}(\bm x, t)}{\alpha_t}, \bm e_d\right\rangle  \bm e_d \\
     & = \sum_{j=1}^{d-1} \theta_j^\ell x_j \bm e_j +
\left(\theta_d^\ell-\frac{\gamma}{\alpha_t}\left(1 - \sigma_t \theta_d^\ell\right)
\right) x_d \bm e_d.
\end{align*}
Then, the last component is
\begin{align*}
   \theta^{\ell,{\rm g}}_{d} := \left( 1 + \frac{\gamma \sigma_t}{\alpha_t} \right)\theta^\ell_{d} - \frac{\gamma}{\alpha_t}.
\end{align*}
\end{proof}

This theorem shows that, for vanilla diffusion models, the learning error on the constrained coordinate remains large in gradient descent, leading to slow correction during sampling. Although guidance amplifies the update on the constrained dimension to reduce the sampling error caused by imperfect training of the score function, achieving \(\theta_d^{\mathrm{g}} \approx \theta_d^*=-\sigma_t^{-2}\) at small \(\sigma_t\) typically requires large \(\gamma\), which is not reasonable in practice because it may introduce other instability issues. Moreover, this also requires very precise choices of $\gamma$ for different inference times $t$, which may require substantially more intensive hyperparameter tuning.

\paragraph{Post-hoc distillation}
Let \(\widehat{\bm{x}}_0(\bm{\epsilon})\) denote the teacher sample obtained by running an ODE solver (probability flow ODE) starting from noise \(\bm{\epsilon}\sim\mathcal{N}(\bm{0},\bm{I})\) using a trained score model (with or without guidance).
We distill this teacher into a student mapping
\[
\bm{v}_{\bm{w}}(\bm{\epsilon}) = \mathrm{diag}(\bm{w})\bm{\epsilon},
\]
trained with a squared loss plus a hard-constraint penalty on the final coordinate:
\[
\mathcal{L}_{\mathrm{distill}}(\bm{w})
=
\mathbb{E}_{\bm{\epsilon}}
\Bigl[
\bigl\|\bm{v}_{\bm{w}}(\bm{\epsilon})-\widehat{\bm{x}}_0(\bm{\epsilon})\bigr\|_2^2
\Bigr]
+
\lambda\,
\mathbb{E}_{\bm{\epsilon}}
\Bigl[
\langle \bm{v}_{\bm{w}}(\bm{\epsilon}),\bm{e}_d\rangle^2
\Bigr],
\qquad \lambda\ge 0.
\]
Then, the following theorem characterizes the solution of the distilled student learner and proves the constraint error achieved by the post-hoc distillation.
\begin{theorem}
\label{thm:distill_closed_form}
Assume \(\bm{\epsilon}\sim\mathcal{N}(\bm{0},\bm{I})\) and \(\bm{v}_{\bm{w}}(\bm{\epsilon})=\mathrm{diag}(\bm{w})\bm{\epsilon}\).
Let \(\widehat{\bm{x}}_0(\bm{\epsilon})\) be any teacher output with \(\mathbb{E}\|\widehat{\bm{x}}_0(\bm{\epsilon})\|_2^2<\infty\).
Then \(\mathcal{L}_{\mathrm{distill}}(\bm{w})\) is strictly convex and its unique minimizer satisfies, for each coordinate \(j\),
\[
w_j^* =
\frac{\mathbb{E}\bigl[\epsilon_j\,\widehat{x}_{0,j}(\bm{\epsilon})\bigr]}{1+\lambda\,\mathbf{1}\{j=d\}}.
\]
In particular, the constrained coordinate is shrunk by a factor \(1/(1+\lambda)\):
\[
w_d^* = \frac{1}{1+\lambda}\,\mathbb{E}\bigl[\epsilon_d\,\widehat{x}_{0,d}(\bm{\epsilon})\bigr],
\qquad
\mathbb{E}\bigl[\langle \bm{v}_{\bm{w}^*}(\bm{\epsilon}),\bm{e}_d\rangle^2\bigr]
=
(w_d^*)^2
\le \frac{1}{(1+\lambda)^2}\,
\mathbb{E}\bigl[\widehat{x}_{0,d}(\bm{\epsilon})^2\bigr].
\]
\end{theorem}

\begin{proof}
Expand the loss coordinate-wise. Since \(\bm{v}_{\bm{w}}(\bm{\epsilon})_j = w_j\epsilon_j\),
\[
\mathcal{L}_{\mathrm{distill}}(\bm{w})
=
\sum_{j=1}^d
\mathbb{E}\bigl[(w_j\epsilon_j-\widehat{x}_{0,j})^2\bigr]
+
\lambda\,\mathbb{E}\bigl[(w_d\epsilon_d)^2\bigr].
\]
Using \(\mathbb{E}[\epsilon_j^2]=1\), we get
\[
\mathbb{E}\bigl[(w_j\epsilon_j-\widehat{x}_{0,j})^2\bigr]
=
w_j^2 - 2w_j\,\mathbb{E}[\epsilon_j\widehat{x}_{0,j}] + \mathbb{E}[\widehat{x}_{0,j}^2].
\]
Thus, for \(j\neq d\), the derivative is \(2w_j-2\mathbb{E}[\epsilon_j\widehat{x}_{0,j}]\), giving
\(w_j^*=\mathbb{E}[\epsilon_j\widehat{x}_{0,j}]\).
For \(j=d\), the coefficient of \(w_d^2\) becomes \(1+\lambda\), giving
\(w_d^*=\mathbb{E}[\epsilon_d\widehat{x}_{0,d}]/(1+\lambda)\).
The bound on the constraint second moment follows by direct substitution.
\end{proof}

\paragraph{Remark.} Based on the above theorem, we can see that even if the teacher \(\widehat{\bm{x}}_0(\bm{\epsilon})\) has a nonzero constraint violation in the \(d\)-th coordinate due to imperfect score learning and/or numerical solver error, the distilled student, through the proposed distillation loss function, can systematically reduce this violation through the penalty parameter \(\lambda\). Moreover, the penalization is applied only along one constrained direction, so learning on the remaining directions is still maintained, and the overall generation quality need not be degraded.


\section{Datasets}\label{appendix:datasets}
We consider the following widely used PDE benchmarks. Each dataset contains paired solution and coefficient fields defined on a $128 \times 128$ grid. These datasets are readily accessible from FNO \cite{li2020fno}, DiffusionPDE \cite{huang2024diffusionpde}, and ECI-Sampling \cite{cheng2025eci}.
\subsection{Darcy Flow}
We adopt the Darcy Flow setup introduced in DiffusionPDE~\cite{huang2024diffusionpde}, with the dataset released by FNO \cite{li2020fno}. For completeness, we describe the generation process here. Specifically, we consider the steady-state Darcy flow equation on a 2D rectangular domain $\Omega \subset \mathbb{R}^2$ with no-slip boundary conditions:
\[
- \nabla \cdot (a(c) \nabla u(c)) = q(c), \quad c \in \Omega, \quad u(c) = 0, \quad c \in \partial \Omega.
\]
Here, $a(c)$ is the spatially varying permeability field with binary values, and $q(c)$ is set to 1 for constant forcing. The pair $(u, a)$ is jointly modeled by the diffusion model.

\subsection{Inhomogeneous Helmholtz Equation and Poisson Equation}
We adopt the setup introduced in DiffusionPDE~\cite{huang2024diffusionpde}, with the dataset released by FNO \cite{li2020fno}. For completeness, we describe the generation process here. As a special case of the inhomogeneous Helmholtz equation, the Poisson equation is obtained by setting $k = 0$:
\[
\nabla^2 u(c) = a(c), \quad c \in \Omega, \quad u(c) = 0, \quad c \in \partial \Omega.
\]
Here, $a(c)$ is a piecewise constant forcing function. The pair $(u, a)$ is jointly modeled by the diffusion model.

\subsection{Burgers' Equation}
We adopt the Burgers' Equation setup introduced in DiffusionPDE~\cite{huang2024diffusionpde}, with the dataset released by FNO \cite{li2020fno}. For completeness, we describe the generation process here. We study the 1D viscous Burgers’ equation with periodic boundary conditions on a spatial domain $\Omega = (0,1)$ and temporal domain $\tau \in (0, T]$:
\[
\partial_\tau u(c, \tau) + \partial_c \left(\frac{u^2(c,\tau)}{2} \right) = \nu \partial^2_{cc} u(c, \tau), \quad u(c, 0) = a(c), \quad c \in \Omega.
\]
In our experiments, we set $\nu = 0.01$. Specifically, we use 128 temporal steps, where each trajectory has shape $128 \times 128$. The pair $(u, a)$ is jointly modeled by the diffusion model.

\subsection{Stokes Problem}
We adopt the Stokes problem setup introduced in ECI-Sampling~\cite{cheng2025eci} and use their released generation code. For completeness, we describe the generation process below.

The 1D Stokes problem is governed by the heat equation:
\[
u_t = \nu u_{xx}, \quad x \in [0, 1],\ t \in [0, 1],
\]
with the following boundary and initial conditions:
\[
u(x, 0) = Ae^{-kx} \cos(kx), \quad x \in [0,1], \quad u(0, t) = A \cos(\omega t), \quad t \in [0,1],
\]
where $\nu \geq 0$ is the viscosity, $A > 0$ is the amplitude, $\omega$ is the oscillation frequency, and $k = \sqrt{\omega / (2 \nu)}$ controls the spatial decay. The analytical solution is given by:
\[
u_{\text{exact}}(x, t) = Ae^{-kx} \cos(kx - \omega t).
\]

In our experiments, we set $A = 2$, $k = 5$ and take $a:=\omega \sim \mathcal{U}[2, 8]$ as the coefficient field to jointly model with $u$.

\subsection{Heat Equation}
We adopt the heat equation setup introduced in ECI-Sampling~\cite{cheng2025eci} and use their released generation code. For completeness, we describe the generation process below.

The 1D heat (diffusion) equation with periodic boundary conditions is defined as:
\[
u_t = \alpha u_{xx}, \quad x \in [0, 2\pi],\ t \in [0, 1],
\]
with the initial and boundary conditions:
\[
u(x, 0) = \sin(x + \varphi), \quad u(0, t) = u(2\pi, t).
\]
Here, $\alpha$ denotes the diffusion coefficient and $\varphi$ controls the phase of the sinusoidal initial condition. The exact solution is:
\[
u_{\text{exact}}(x, t) = e^{-\alpha t} \sin(x + \varphi).
\]

In our experiments, we set $\alpha=3$ and take $a:=\varphi \sim \mathcal{U}[0, \pi]$ as the coefficient to jointly model with $u$.

\subsection{Navier--Stokes Equation}
We adopt the 2D Navier--Stokes (NS) setup from ECI-Sampling~\cite{cheng2025eci} and use their released generation code. The NS equation in vorticity form for an incompressible fluid with periodic boundary conditions is given as:
\begin{align*}
\partial_t w(x,t) + u(x,t) \cdot \nabla w(x,t) &= \nu \Delta w(x,t) + f(x), \quad x \in [0,1]^2, \, t \in [0,T], \\
\nabla \cdot u(x,t) &= 0, \quad x \in [0,1]^2, \, t \in [0,T], \\
w(x,0) &= w_0(x), \quad x \in [0,1]^2.
\end{align*}
Here, $u$ denotes the velocity field and $w = \nabla \times u$ is the vorticity. The initial vorticity $w_0$ is sampled from $\mathcal{N}(0, 7^{3/2}(-\Delta + 49I)^{-5/2})$, and the forcing term is defined as $f(x) = 0.1 \sqrt{2} \sin(2\pi(x_1 + x_2) + \phi)$, where $\phi \sim \mathcal{U}[0, \pi/2]$. We take $a:=w_0$ as the coefficient to jointly model with $u$.

\subsection{Porous Medium Equation}
We use the Porous Medium Equation (PME) setup provided by ECI-Sampling~\cite{cheng2025eci}, with zero initial and time-varying Dirichlet left boundary conditions:
\begin{align*}
u_t &= \nabla \cdot (u^m \nabla u), \quad x \in [0,1], \, t \in [0,1], \\
u(x,0) &= 0, \quad x \in [0,1], \\
u(0,t) &= (mt)^{1/m}, \quad t \in [0,1], \\
u(1,t) &= 0, \quad t \in [0,1].
\end{align*}
The exact solution is $u_{\text{exact}}(x,t) = (m \cdot \text{ReLU}(t - x))^{1/m}$. The exponent $m$ is sampled from $\mathcal{U}[1,5]$. We take $a:=m$ as the coefficient to jointly model with $u$.

\subsection{Stefan Problem}
We also adopt the Stefan problem configuration from ECI-Sampling~\cite{cheng2025eci}, which is a nonlinear case of the Generalized Porous Medium Equation (GPME) with fixed Dirichlet boundary conditions:
\begin{align*}
u_t &= \nabla \cdot (k(u)\nabla u), \quad x \in [0,1], \, t \in [0,T], \\
u(x,0) &= 0, \quad x \in [0,1], \\
u(0,t) &= 1, \quad t \in [0,T], \\
u(1,t) &= 0, \quad t \in [0,T],
\end{align*}
where $k(u)$ is a step function defined by a shock value $u^*$:
\begin{equation*}
k(u) =
\begin{cases}
1, & u \geq u^*, \\
0, & u < u^*.
\end{cases}
\end{equation*}
The exact solution is:
\begin{equation*}
u_{\text{exact}}(x,t) = \mathbb{1}_{[u \geq u^*]} \left(1 - (1 - u^*) \frac{\operatorname{erf}(x/(2\sqrt{t}))}{\operatorname{erf}(\alpha)}\right),
\end{equation*}
where $\alpha$ satisfies the nonlinear equation $(1 - u^*)/\sqrt{\pi} = u^* \operatorname{erf}(\alpha)\alpha\exp(\alpha^2)$. We follow ECI-Sampling to take $a:=u^* \sim \mathcal{U}[0.55, 0.7]$ as the coefficient to jointly model with $u$.

\section{Experimental Setup}\label{appendix:setup}

This section provides details on the model architecture, training configurations for diffusion and distillation, evaluation protocols, and baseline methods.

\subsection{Model Structure}

We follow ECI-sampling \cite{cheng2025eci} and adopt the Fourier Neural Operator (FNO) \cite{li2020fno} as both the teacher diffusion model and the student distillation model. A sinusoidal positional encoding \cite{vaswani2017attention} is appended as an additional input dimension. Specifically, we use a four-layer FNO with a frequency cutoff of $32 \times 32$, a time embedding dimension of 32, a hidden channel width of 64, and a projection dimension of 256.

\subsection{Diffusion and Distillation Training Setup}

For diffusion training, we employ a standard linear noise schedule~\cite{liu2023rectifiedflow, liu2023flow, lipman2023flow, liu2023instaflow} with a batch size of 128 and a total of 10,000 iterations. The model is optimized using Adam \cite{kingma2014adam} with a learning rate of $3 \times 10^{-2}$.

During distillation, we use Euler’s method with 100 uniformly spaced timesteps from $t = 1$ to $t = 0$ for sampling. Every 100 epochs, we resample 1024 new noise–data pairs for supervision. Distillation is trained for 2000 epochs using Adam (learning rate $3 \times 10^{-2}$), with early stopping based on the squared norm of the observation loss, i.e., $\lVert d_{\theta'}(\varepsilon) - x \rVert^2$.

The physics constraint weight $\lambda_{\text{train}}$ is set to 10 for Darcy Flow, Burgers' Equation, Stokes Problem, Heat Equation, Navier–Stokes, Porous Medium Equation, and Stefan Problem. For Helmholtz and Poisson equations, we increase $\lambda_{\text{train}}$ to $10^6$ due to the stiffness of these PDEs. All experiments are conducted on an NVIDIA RTX 4090 GPU.

\subsection{Evaluation Setup}

For physics-based data simulation, we evaluate models with and without physics refinement: the number of gradient-based refinement steps $N$ is set to 0 or 50. The step size $\eta$ is aligned with the dataset-specific $\lambda_{\text{train}}$ used during distillation.

In forward and inverse problems, the observation mask $M$ defines the known entries. For forward problems, the mask has ones at boundary entries. For partial reconstruction, the mask is sampled randomly with 20\% of entries set to 1 (observed), and the rest to 0 (missing). All evaluations are conducted on an NVIDIA RTX 4090 GPU.

\subsection{Baseline Methods}\label{appendix:baseline}

We describe the configurations of all baseline methods used for comparison. Where necessary, we adapt our diffusion training and sampling codebase to implement their respective constraint mechanisms.

\textbf{ECI-sampling.} We follow the approach of directly substituting hard constraints into the posterior mean $\mathbb{E}[x_0 \mid x_t]$ based on a predefined observation mask. Specifically, we project these constraints at each DDIM step \cite{song2021ddim} using a correction operator $C$:
\begin{equation}
\vx_{t - dt} = C(\hat{\vx}_\theta(\vx_t, t)) \cdot (1 - t + \mathrm{d}t) + (\vx_t - \hat{\vx}_\theta(\vx_t, t)) \cdot (t - \mathrm{d}t),
\end{equation}
where $t$ flows backward from 1 to 0, and $\hat{\vx}_\theta$ denotes the posterior mean estimated using Tweedie's formula.

\textbf{DiffusionPDE.} This method employs diffusion posterior sampling (DPS) \cite{chung2022dps}, where each intermediate sample $\vx_t$ is guided by the gradient of the PDE residual evaluated on the posterior mean:
\begin{equation}
\vx_{t - dt} = \vx_t + v_{\theta}(\vx_t, t) \cdot \mathrm{d}t - \eta_t \nabla_{\vx_t} \left\| \mathcal{R}(\hat{\vx}_\theta(\vx_t, t)) \right\|^2,
\end{equation}
where $v_{\theta}(\vx_t, t)$ is the learned velocity field from the reverse-time ODE sampler, and $\eta_t$ is a hyperparameter. In our experiments, we set $\eta_t$ equal to $\lambda_{\text{train}}$ for each dataset.

\textbf{PIDM.} This method incorporates an additional residual loss into the diffusion training objective, evaluated on the posterior mean $\mathbb{E}[x_0 \mid x_t]$. Specifically, PIDM~\cite{bastek2025pdim} augments the standard diffusion loss with a physics-based term:
\begin{equation}
\mathcal{L}_{\text{PIDM}} = \mathcal{L}_{\text{diffusion}} + \lambda_t \left\| \mathcal{R}(\hat{\vx}_\theta(\vx_t, t)) \right\|^2,
\end{equation}
where $\mathcal{L}_{\text{diffusion}}$ is the original diffusion training loss, and $\lambda_t$ is the residual loss weight. In our experiments, we set $\lambda_t$ to $\lambda_{\text{train}}$ for each dataset as it gives stable performance.

\textbf{D-Flow.} For this standard method \cite{ben2024dflow}, we build on the official implementation of ECI-sampling \cite{cheng2025eci} and introduce an additional PDE residual loss evaluated on the final sample. The weighting $\lambda_{\text{train}}$ is aligned with our setup across datasets. Specifically, the implementation follows the D-Flow setup in ECI-sampling \cite{cheng2025eci}: we discretize the sampling trajectory into 100 denoising steps and perform gradient-based optimization on the input noise over 50 iterations to minimize the physics residual loss. At each iteration, gradients are backpropagated through the entire 100-step trajectory, resulting in a total of 5,000 function evaluations (NFE) per sample. This leads to significantly higher computational cost compared to our one-step method.

\textbf{Teacher.} This baseline refers to sampling directly from the trained teacher diffusion model without incorporating any PDE-based constraint or guidance mechanism.

\section{Generative Evaluations on More Datasets}\label{appendix:results_more_data}
In this section, we report results on more datasets and compare them with other baseline methods, as shown in Table~\ref{tab:pde_results_more_data}. PIDDM consistently improves over the baselines, especially in physics residual error.
\begin{table*}[ht]
\centering
\small
\setlength{\tabcolsep}{3pt}
\renewcommand{\arraystretch}{1.2}
\caption{Generative metrics on various constrained PDEs. PDE error is the MSE of the evaluated physics residual error. The best results are in \textbf{bold}.}
\label{tab:pde_results_more_data}
\begin{tabular}{ll
                S  S  S  S  S  S  S}
\toprule
Dataset & Metric &
  {PIDDM-1} & {PIDDM-ref} & {ECI} & {DiffusionPDE} &
  {D-Flow} & {PIDM} & {FM}\\
\midrule
\multirow{4}{*}{Helmholtz}
    & MMSE ($\times 10^{-1}$) &
    {0.265} & {\textbf{0.185}} & {0.318} & {0.335} & {0.140} & {0.352} & {0.296} \\
  & SMSE ($\times 10^{-1}$) &
    {0.195} & {\textbf{0.169}} & {0.289} & {0.301} & {0.106} & {0.325} & {0.210} \\
  & PDE Error ($\times 10^{-9}$) &
    {0.054} & {\textbf{0.034}} & {2.135} & {1.812} & {0.680} & {1.142} & {2.104} \\
  & NFE ($\times 10^{3}$) &
    {\textbf{0.001}} & {0.100} & {0.500} & {0.100} & {5.000} & {0.100} & {0.100} \\
\midrule
\multirow{4}{*}{Stokes Problem}
  & MMSE ($\times 10^{-2}$) &
    {0.298} & {\bfseries 0.182} & {0.335} & {0.342} & {0.301} & {0.361} & {0.310} \\
  & SMSE ($\times 10^{-2}$) &
    {0.425} & {\bfseries 0.312} & {0.455} & {0.469} & {0.441} & {0.484} & {0.430} \\
  & PDE Error ($\times 10^{-3}$) &
    {0.241} & {\bfseries 0.194} & {0.585} & {0.498} & {0.318} & {0.432} & {0.578} \\
  & NFE ($\times 10^{3}$) &
    {\bfseries 0.001} & {0.100} & {0.500} & {0.100} & {5.000} & {0.100} & {0.100} \\
\midrule
\multirow{4}{*}{Heat Equation}
  & MMSE ($\times 10^{-3}$) &
    {0.901} & {\bfseries 0.845} & {4.620} & {4.600} & {1.452} & {4.580} & {4.544} \\
  & SMSE ($\times 10^{-2}$) &
    {0.816} & {\bfseries 0.790} & {1.612} & {1.598} & {0.892} & {1.587} & {1.565} \\
  & PDE Error ($\times 10^{-5}$) &
    {3.265} & {\bfseries 2.910} & {4.120} & {4.100} & {3.698} & {4.150} & {4.354} \\
  & NFE ($\times 10^{3}$) &
    {\bfseries 0.001} & {0.100} & {0.500} & {0.100} & {5.000} & {0.100} & {0.100} \\
\midrule
\multirow{4}{*}{\makecell{Navier–\\Stokes\\ Equation}}
  & MMSE ($\times 10^{-4}$) &
    {0.285} & {\bfseries 0.264} & {0.302} & {0.299} & {0.288} & {0.306} & {0.294} \\
  & SMSE ($\times 10^{-4}$) &
    {0.218} & {\bfseries 0.210} & {0.323} & {0.321} & {0.225} & {0.327} & {0.314} \\
  & PDE Error ($\times 10^{-5}$) &
    {3.184} & {\bfseries 2.945} & {6.910} & {6.740} & {3.200} & {6.950} & {7.222} \\
  & NFE ($\times 10^{3}$) &
    {\bfseries 0.001} & {0.100} & {0.500} & {0.100} & {5.000} & {0.100} & {0.100} \\
\midrule
\multirow{4}{*}{\makecell{Porous\\ Medium\\ Equation}}
  & MMSE ($\times 10^{-3}$) &
    {4.555} & {\bfseries 4.210} & {7.742} & {7.698} & {5.203} & {7.762} & {7.863} \\
  & SMSE ($\times 10^{-1}$) &
    {2.143} & {\bfseries 2.051} & {2.573} & {2.602} & {2.327} & {2.589} & {2.639} \\
  & PDE Error ($\times 10^{-5}$) &
    {3.412} & {\bfseries 3.110} & {4.982} & {4.945} & {3.548} & {4.917} & {5.523} \\
  & NFE ($\times 10^{3}$) &
    {\bfseries 0.001} & {0.100} & {0.500} & {0.100} & {5.000} & {0.100} & {0.100} \\
\midrule
\multirow{4}{*}{Stefan Problem}
  & MMSE ($\times 10^{-3}$) &
    {0.231} & {\bfseries 0.220} & {0.248} & {0.249} & {0.238} & {0.252} & {0.245} \\
  & SMSE ($\times 10^{-3}$) &
    {0.278} & {\bfseries 0.268} & {0.315} & {0.318} & {0.289} & {0.320} & {0.307} \\
  & PDE Error ($\times 10^{-2}$) &
    {0.081} & {\bfseries 0.070} & {0.410} & {0.398} & {0.095} & {0.405} & {0.458} \\
  & NFE ($\times 10^{3}$) &
    {\bfseries 0.001} & {0.100} & {0.500} & {0.100} & {5.000} & {0.100} & {0.100} \\
\bottomrule
\end{tabular}
\end{table*}

\FloatBarrier
\begin{figure*}[ht]
\centering
\begin{subfigure}{0.30\linewidth}
    \centering
    \includegraphics[width=\linewidth]{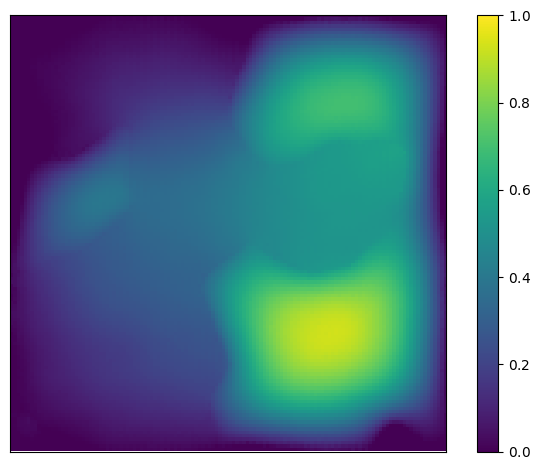}
    \caption*{ECI-sampling – Solution}
\end{subfigure}
\begin{subfigure}{0.30\linewidth}
    \centering
    \includegraphics[width=\linewidth]{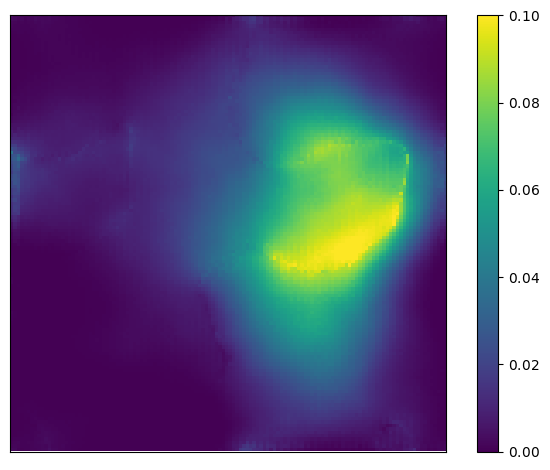}
    \caption*{ECI-sampling – Data Error}
\end{subfigure}
\begin{subfigure}{0.30\linewidth}
    \centering
    \includegraphics[width=\linewidth]{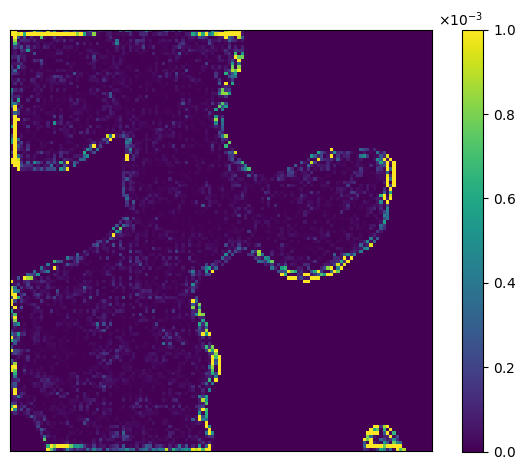}
    \caption*{ECI-sampling – PDE Error}
\end{subfigure}\\[2pt]

\begin{subfigure}{0.30\linewidth}
    \centering
    \includegraphics[width=\linewidth]{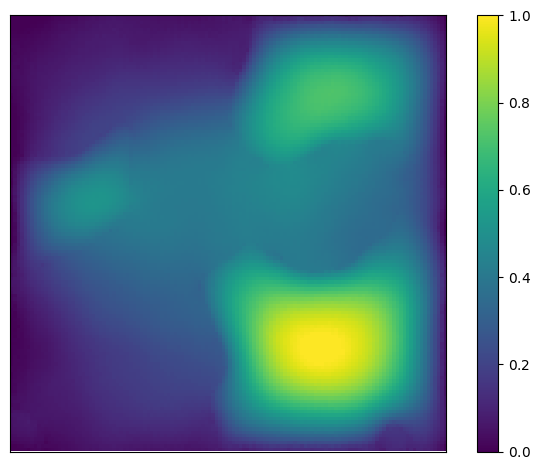}
    \caption*{DiffusionPDE – Solution}
\end{subfigure}
\begin{subfigure}{0.30\linewidth}
    \centering
    \includegraphics[width=\linewidth]{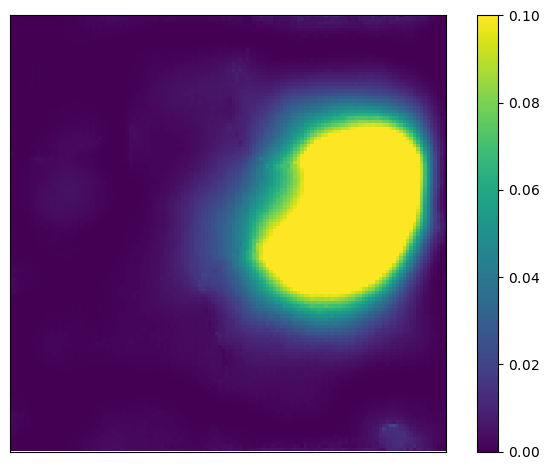}
    \caption*{DiffusionPDE – Data Error}
\end{subfigure}
\begin{subfigure}{0.30\linewidth}
    \centering
    \includegraphics[width=\linewidth]{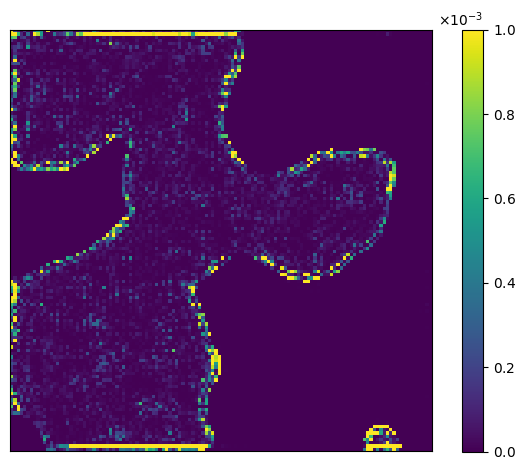}
    \caption*{DiffusionPDE – PDE Error}
\end{subfigure}\\[2pt]

\begin{subfigure}{0.30\linewidth}
    \centering
    \includegraphics[width=\linewidth]{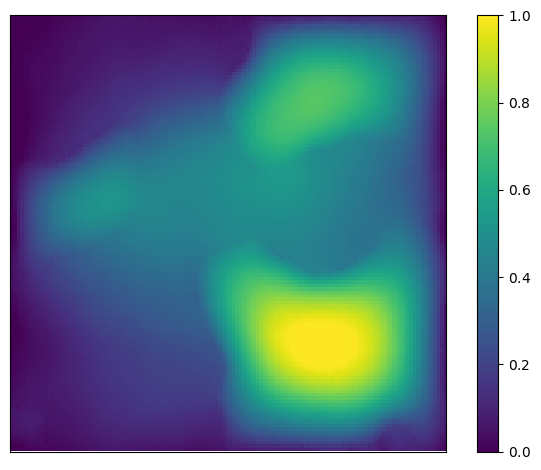}
    \caption*{D-Flow – Solution}
\end{subfigure}
\begin{subfigure}{0.30\linewidth}
    \centering
    \includegraphics[width=\linewidth]{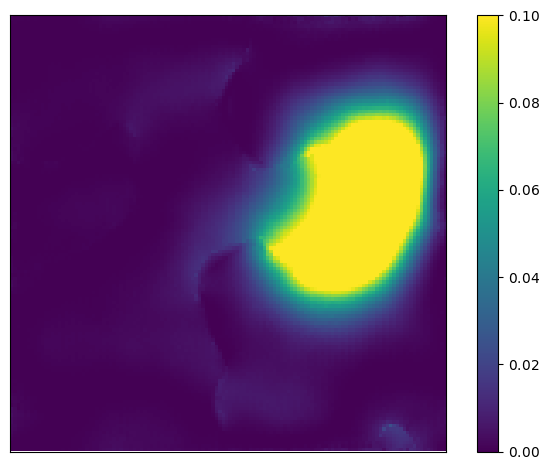}
    \caption*{D-Flow – Data Error}
\end{subfigure}
\begin{subfigure}{0.30\linewidth}
    \centering
    \includegraphics[width=\linewidth]{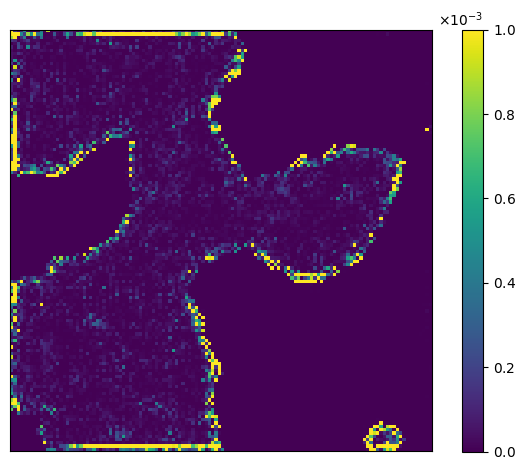}
    \caption*{D-Flow – PDE Error}
\end{subfigure}\\[2pt]

\begin{subfigure}{0.30\linewidth}
    \centering
    \includegraphics[width=\linewidth]{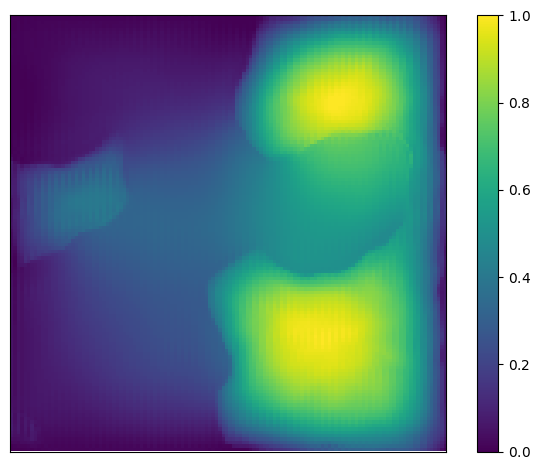}
    \caption*{PIDDM – Solution}
\end{subfigure}
\begin{subfigure}{0.30\linewidth}
    \centering
    \includegraphics[width=\linewidth]{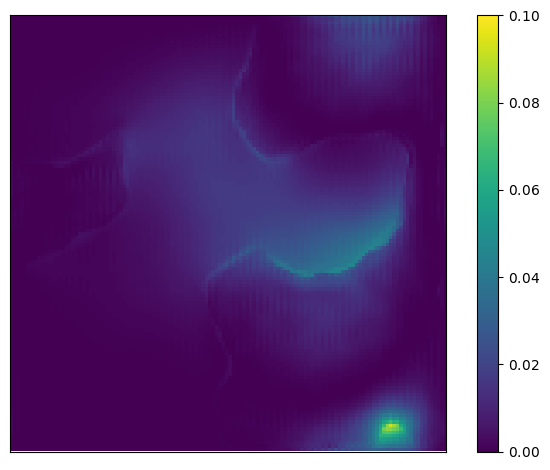}
    \caption*{PIDDM – Data Error}
\end{subfigure}
\begin{subfigure}{0.30\linewidth}
    \centering
    \includegraphics[width=\linewidth]{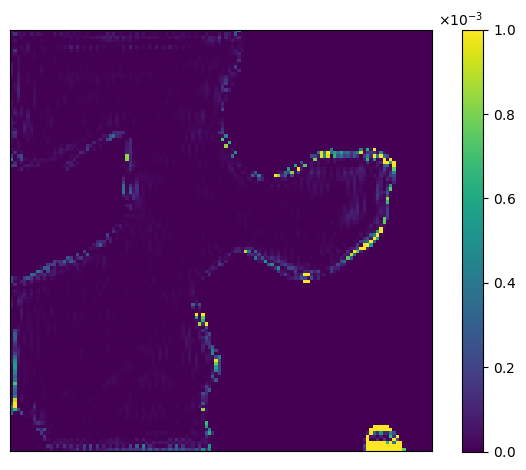}
    \caption*{PIDDM – PDE Error}
\end{subfigure}

\caption{Qualitative comparison on the Darcy \emph{forward} problem.
Each column shows (left) the predicted solution field, (middle) pointwise data error, and (right) PDE residual error.
Our PIDDM (bottom row) delivers visibly lower data and PDE errors than other baselines while maintaining sharp solution details.}
\label{fig:vis_error_darcy_forward}
\end{figure*}

\section{Qualitative Results on the Darcy Forward Problem}
Figure~\ref{fig:vis_error_darcy_forward} compares the predicted Darcy pressure fields and their corresponding data- and PDE-error maps for each baseline and for our PIDDM. DiffusionPDE and ECI reproduce the coarse flow pattern but exhibit large pointwise errors and pronounced residual bands. In contrast, PIDDM produces the visually sharpest solution and the lowest error intensities in both maps, confirming the quantitative gains reported in the main text.
\FloatBarrier
\section{Additional Experiments}\label{appendix:additional_experiments}

\subsection{Additional Review-Driven Ablations}

We include additional experiments requested during review to clarify component attribution, diversity preservation, teacher-quality sensitivity, architecture robustness, dynamic loss weighting, and latent-optimization stability. These experiments use the same evaluation protocol as the main text and are reported on Darcy unless otherwise specified. The compute-matched comparison is reported in Table~\ref{tab:compute_matched}.

\paragraph{Component attribution.} Table~\ref{tab:appendix_component_attribution} separates the effects of one-step distillation and final-sample PDE supervision. Distillation alone mainly provides the efficiency gain but does not improve PDE satisfaction, whereas final-sample PDE supervision is the key component for physical consistency. PIDDM combines both effects and achieves a better quality--compute trade-off.

\paragraph{Diversity preservation.} Beyond MMSE, SMSE, and FPD, we report sliced Wasserstein distance (SWD) and average pairwise distance difference (APDD). For samples \(x^{(1)},\ldots,x^{(N)}\), define
\[
\mathrm{APD}=\frac{2}{N(N-1)}\sum_{i<j}\|x_i-x_j\|_2,
\qquad
\mathrm{APDD}=\frac{\left|\mathrm{APD}_S-\mathrm{APD}_T\right|}{\mathrm{APD}_T},
\]
where \(\mathrm{APD}_S\) and \(\mathrm{APD}_T\) are computed on student and teacher samples, respectively. Table~\ref{tab:appendix_diversity} shows that PIDDM preserves diversity better than vanilla distillation, with PIDDM-ref performing best.

\paragraph{Teacher quality and architecture robustness.} Table~\ref{tab:appendix_teacher_quality} shows that student performance improves with teacher checkpoint quality early in training, but gains become marginal after 6k teacher iterations. Table~\ref{tab:appendix_architecture} further verifies that PIDDM improves physical consistency over DiffusionPDE under U-Net and CNN auto-encoder backbones, showing that the final-sample supervision strategy is not tied to FNO.

\paragraph{Dynamic weighting and latent optimization stability.} For stiff systems such as Helmholtz and Poisson, raw PDE residual magnitudes can differ by orders of magnitude across PDE families. We therefore evaluate a dynamic weighting variant that rescales the PDE term by a detached running residual magnitude to target a comparable scale of \(10^{-4}\). Table~\ref{tab:appendix_dynamic_weighting} shows comparable performance to manually tuned weights. Tables~\ref{tab:appendix_latent_convergence} and~\ref{tab:appendix_latent_variance} show that the downstream latent optimization converges quickly and has lower variance across random initializations than DiffusionPDE.

\paragraph{Final-sample PDE supervision across distillation backbones.} Table~\ref{tab:appendix_final_sample_supervision} compares standard distillation backbones with and without final-sample PDE supervision (PS). Removing PS only slightly changes MSE but increases PDE error by about one order of magnitude, indicating that the backbone mainly affects one-step distillation quality while final-sample PDE supervision drives the physics-consistency gain.

\begin{table}[!htbp]
\centering
\setlength{\tabcolsep}{4pt}
\renewcommand{\arraystretch}{1.15}
\caption{Component attribution on Darcy generation. Distill denotes pure one-step distillation without final-sample PDE supervision.}
\label{tab:appendix_component_attribution}
\begin{tabular}{lrrrr}
\toprule
Metric & PIDDM-ref & Distill & Teacher & D-Flow \\
\midrule
MMSE ($\times 10^{-2}$) & 0.037 & 0.121 & 0.108 & 0.129 \\
SMSE ($\times 10^{-2}$) & 0.002 & 0.102 & 0.069 & 0.085 \\
PDE error ($\times 10^{-4}$) & 0.148 & 1.642 & 1.585 & 0.532 \\
FPD & 0.385 & 0.825 & 0.782 & 0.995 \\
NFE & 80 & 1 & 100 & 5000 \\
\bottomrule
\end{tabular}
\end{table}

\begin{table}[!htbp]
\centering
\setlength{\tabcolsep}{4pt}
\renewcommand{\arraystretch}{1.15}
\caption{Diversity preservation on Darcy generation. Lower SWD and APDD indicate closer distributional diversity to the reference samples.}
\label{tab:appendix_diversity}
\begin{tabular}{lrrrr}
\toprule
Metric & PIDDM-1 & PIDDM-ref & Distillation & Teacher \\
\midrule
SWD & 0.058 & 0.032 & 0.103 & 0.089 \\
APDD & 0.050 & 0.032 & 0.067 & 0.034 \\
\bottomrule
\end{tabular}
\end{table}

\begin{table}[!htbp]
\centering
\setlength{\tabcolsep}{4pt}
\renewcommand{\arraystretch}{1.15}
\caption{Sensitivity to teacher checkpoint quality on Darcy generation.}
\label{tab:appendix_teacher_quality}
\begin{tabular}{lrrrrr}
\toprule
Metric & 2k & 4k & 6k & 8k & 10k \\
\midrule
MMSE ($\times 10^{-2}$) & 0.174 & 0.136 & 0.121 & 0.119 & 0.112 \\
SMSE ($\times 10^{-2}$) & 0.124 & 0.098 & 0.085 & 0.085 & 0.082 \\
PDE error ($\times 10^{-4}$) & 0.336 & 0.247 & 0.231 & 0.229 & 0.226 \\
\bottomrule
\end{tabular}
\end{table}

\begin{table}[!htbp]
\centering
\setlength{\tabcolsep}{3pt}
\renewcommand{\arraystretch}{1.15}
\caption{Architecture robustness on Darcy generation. PIDDM improves over posterior-mean guidance when using U-Net and CNN auto-encoder backbones.}
\label{tab:appendix_architecture}
\begin{tabular}{lrrrr}
\toprule
Metric & PIDDM(UNet) & DiffusionPDE(UNet) & PIDDM(AE) & DiffusionPDE(AE) \\
\midrule
MMSE ($\times 10^{-2}$) & 0.243 & 0.753 & 0.310 & 0.975 \\
SMSE ($\times 10^{-2}$) & 0.211 & 0.431 & 0.257 & 0.692 \\
PDE error ($\times 10^{-4}$) & 0.743 & 1.566 & 0.902 & 1.848 \\
FPD & 0.948 & 1.651 & 1.182 & 1.906 \\
\bottomrule
\end{tabular}
\end{table}

\begin{table}[!htbp]
\centering
\setlength{\tabcolsep}{3pt}
\renewcommand{\arraystretch}{1.15}
\caption{Dynamic weighting for stiff PDEs. Dynamic weighting rescales the PDE term by a detached running residual magnitude, targeting a comparable residual scale of $10^{-4}$.}
\label{tab:appendix_dynamic_weighting}
\begin{tabular}{lrrrr}
\toprule
Metric & Helmholtz(dynamic) & Helmholtz(manual) & Poisson(dynamic) & Poisson(manual) \\
\midrule
MMSE ($\times 10^{-1}$) & 0.257 & 0.265 & 0.017 & 0.016 \\
SMSE ($\times 10^{-1}$) & 0.184 & 0.195 & 0.035 & 0.036 \\
PDE error ($\times 10^{-9}$) & 0.053 & 0.054 & 0.069 & 0.073 \\
\bottomrule
\end{tabular}
\end{table}

\begin{table}[!htbp]
\centering
\setlength{\tabcolsep}{5pt}
\renewcommand{\arraystretch}{1.15}
\caption{Convergence of latent optimization on the Darcy forward task.}
\label{tab:appendix_latent_convergence}
\begin{tabular}{lrrrrr}
\toprule
Metric & 10 & 20 & 40 & 60 & 80 \\
\midrule
PDE loss ($\times 10^{-4}$) & 0.727 & 0.173 & 0.152 & 0.149 & 0.145 \\
Data MSE & 0.119 & 0.012 & 0.003 & 0.002 & 0.002 \\
\bottomrule
\end{tabular}
\end{table}

\begin{table}[!htbp]
\centering
\setlength{\tabcolsep}{5pt}
\renewcommand{\arraystretch}{1.15}
\caption{Error variance across random initializations. Values are MSE variance ($\times 10^{-5}$).}
\label{tab:appendix_latent_variance}
\begin{tabular}{lrrr}
\toprule
Method & Forward & Inverse & Reconstruction \\
\midrule
PIDDM-ref & 8.57 & 9.49 & 8.14 \\
DiffusionPDE & 46.39 & 54.61 & 42.72 \\
\bottomrule
\end{tabular}
\end{table}

\begin{table*}[!htbp]
\scriptsize
\centering
\setlength{\tabcolsep}{3pt}
\renewcommand{\arraystretch}{1.15}
\caption{Effect of final-sample PDE supervision (PS) across one-step distillation backbones on the Darcy forward task.}
\label{tab:appendix_final_sample_supervision}
\begin{tabular}{lrrrrrrrr}
\toprule
Metric & RF-2+PS & RF-2 w/o PS & DMD+PS & DMD w/o PS & CM+PS & CM w/o PS & DiffusionPDE & PIDM \\
\midrule
MSE ($\times 10^{-1}$) & 0.127 & 0.143 & 0.255 & 0.261 & 0.283 & 0.291 & 0.691 & 0.380 \\
PDE error ($\times 10^{-4}$) & 0.098 & 1.304 & 0.134 & 1.248 & 0.083 & 1.373 & 1.576 & 1.248 \\
\bottomrule
\end{tabular}
\end{table*}

\FloatBarrier

\subsection{Correlated MoG Constraint Satisfaction}
\begin{figure*}[ht]
\centering
\begin{subfigure}{0.45\textwidth}
    \includegraphics[width=\linewidth]{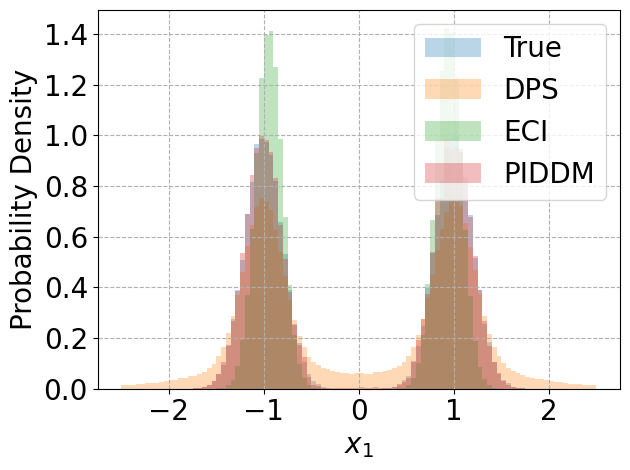}
    \caption{Marginal distribution over $x_1$}
\end{subfigure}
\hfill
\begin{subfigure}{0.45\textwidth}
    \includegraphics[width=\linewidth]{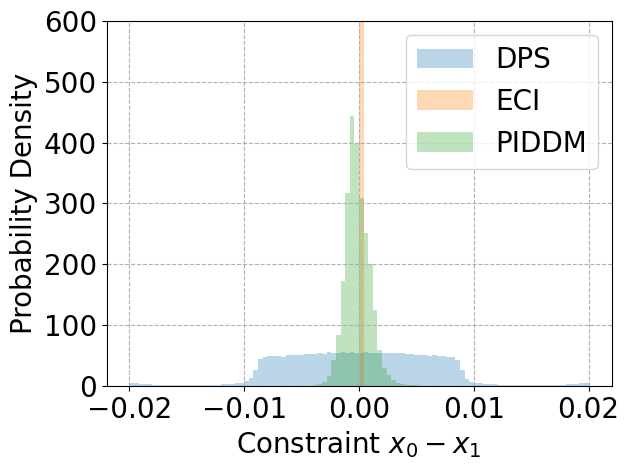}
    \caption{Constraint deviation $x_0 - x_1$}
\end{subfigure}
\vspace{-5pt}
\caption{Constraint satisfaction on correlated MoG. Comparison of generated samples using DPS, ECI, and PIDDM. PIDDM closely matches the target distribution while satisfying constraints.}
\label{fig:mog_highcorr}
\vspace{-10pt}
\end{figure*}

\noindent
We investigate a controlled Mixture-of-Gaussians (MoG) setting to evaluate constraint satisfaction in generative models. The target distribution is a correlated, two-component Gaussian mixture:
\begin{equation}
p_{\mathrm{MoG}}(\mathbf x)=\tfrac{1}{2}\mathcal{N}\left(\mathbf x; [-1,-1]^\top, \Sigma \right) + \tfrac{1}{2}\mathcal{N}\left(\mathbf x; [+1,+1]^\top, \Sigma \right),
\end{equation}
where
\[
\Sigma = \sigma^2
\begin{bmatrix}
1 & \rho \\
\rho & 1
\end{bmatrix},
\qquad
\sigma^2 = 0.04,\quad \rho = 0.99999.
\]
The high correlation $\rho=0.99999$ ensures that the analytic score function $\nabla_{\mathbf{x}} \log p_{\mathrm{MoG}}(\mathbf{x})$ remains well-defined, despite the near-singular covariance. The physical constraint is defined as:
\begin{equation}
\mathcal{F}(\vx) = | \vx_0 - \vx_1 |^2 = 0.
\end{equation}
\textbf{Baselines.} DPS and ECI both integrate the analytical score using 1000-step Euler discretization over $(0, 1)$. DPS applies constraint guidance via gradient descent on $\mathcal{F}(\vx)$ at each step, using a loss weight of $300$. ECI enforces the constraint by directly projecting the posterior mean to satisfy $\mathcal{F}(\vx) = 0$.

\textbf{PIDDM.} A teacher diffusion model is constructed using a probability-flow ODE with 100-step Euler integration, leveraging the analytic score. It generates 50,000 training pairs $(\boldsymbol{\varepsilon}, \vx_0)$ that are used to train a one-step student model, a ReLU-activated MLP with two hidden layers (100 neurons each) via the loss:
\begin{equation}
\mathcal{L}_{\text{train}} = \frac{1}{N} \sum_{i=1}^{N} \left| d_{\boldsymbol{\theta}}(\boldsymbol{\varepsilon}_i) - \boldsymbol{x}_{0,i} \right|^2 + \lambda_{\text{train}} \mathcal{F}(d_{\boldsymbol{\theta}}(\boldsymbol{\varepsilon}_i)), \quad \lambda_{\text{train}} = 1.
\end{equation}
Training uses Adam optimizer (lr = $10^{-3}$, batch size = 2048). During inference, latent noise $\boldsymbol{\varepsilon}$ is optimized via 80 steps of \textsc{LBFGS} with strong-Wolfe line search, learning rate $3 \times 10^{-3}$, and gradient tolerance $10^{-7}$, with $\lambda_{\text{infer}} = 1$.

\textbf{Results.} Figure~\ref{fig:mog_highcorr}(a) shows that all methods recover the bimodal structure of $\vx_1$. However, as shown in Figure~\ref{fig:mog_highcorr}(b), DPS fails to fully satisfy the constraint, with $\vx_0 - \vx_1$ spread over $\pm 10^{-2}$, while ECI enforces it exactly but distorts the marginal distribution. In contrast, PIDDM maintains both constraint satisfaction (standard deviation $\approx 2\times10^{-3}$) and distributional fidelity.


\end{document}